\documentclass[5p]{elsarticle}

\usepackage{hyperref}

\usepackage{lineno}
\usepackage{graphicx}
\usepackage{subfigure}
\usepackage{amsmath,amssymb}
\usepackage{multirow}
\usepackage{algorithm}
\usepackage[]{algorithmic}
\usepackage{siunitx}

\usepackage[T1]{fontenc}
\usepackage[utf8]{inputenc}
\usepackage{soulutf8}

\usepackage{color}
\usepackage{soul}

\usepackage[capitalise, noabbrev]{cleveref}

\journal{Knowledge-Based Systems}
\begin{document}
	
	\begin{frontmatter}
		
% 		\title{Adversarial Domain-invariant Feature Extraction with Transformer and Self-Knowledge Distillation}
% 		\title{Cross-subject adversarial learning for sensor-based human activity recognition with transformer network and self-knowledge distillation}
% 		\title{Cross-subject adversarial learning for human activity recognition with transformer network}
        % \title{Adversarial cross-domain sensor-based human activity recognition with transformer network}
% 		\title{Little strokes fell great oaks: Adversarial cross-domain sensor-based human activity recognition with transformer network and self-knowledge distillation}
		\title{TASKED: Transformer-based Adversarial learning for human activity recognition using wearable sensors via Self-KnowledgE Distillation}
		\author[DFKI,TUKaiserslautern]{Sungho Suh\fnref{cofirst}\corref{mycorrespondingauthor}}
		\cortext[mycorrespondingauthor]{Corresponding author}
		\ead{Sungho.Suh@dfki.de}
		\author[DFKI,TUKaiserslautern]{Vitor Fortes Rey\fnref{cofirst}}
		\author[DFKI,TUKaiserslautern]{Paul Lukowicz}
		
		\fntext[cofirst]{These authors contributed equally to this work.}
		\address[DFKI]{German Research Center for Artificial Intelligence (DFKI), 67663 Kaiserslautern, Germany}
		\address[TUKaiserslautern]{Department of Computer Science, Technische Universität Kaiserslautern, 67663 Kaiserslautern, Germany}

		\begin{abstract}
		    Wearable sensor-based human activity recognition (HAR) has emerged as a principal research area and is utilized in a variety of applications. Recently, deep learning-based methods have achieved significant improvement in the HAR field with the development of human-computer interaction applications. However, most of them are limited to operating in a local neighborhood in the process of a standard convolution neural network, and correlations between different sensors on body positions are ignored. In addition, even though several recent existing works considered the correlations between different sensor positions, they still face significant challenging problems with performance degradation due to large gaps in the distribution of training and test data, and behavioral differences between subjects. In this work, we propose a novel Transformer-based Adversarial learning framework for human activity recognition using wearable sensors via Self-KnowledgE Distillation (TASKED), that accounts for individual sensor orientations and spatial and temporal features. The proposed method is capable of learning cross-domain embedding feature representations from multiple subjects datasets using adversarial learning and the maximum mean discrepancy (MMD) regularization to align the data distribution over multiple domains. In the proposed method, we adopt the teacher-free self-knowledge distillation to improve the stability of the training procedure and the performance of human activity recognition. Experimental results show that TASKED not only outperforms state-of-the-art methods on the four real-world public HAR datasets (alone or combined) but also improves the subject generalization effectively. 
		\end{abstract}
		
		\begin{keyword}
			human activity recognition \sep wearable sensors \sep domain generalization \sep adversarial learning \sep transformer \sep self-knowledge distillation
		\end{keyword}
	\end{frontmatter}
	
	% \linenumbers
	
	\section{Introduction}
	\label{introduction}
	Wearable sensor-based human activity recognition (HAR) has become an important field with various applications, such as health management \cite{bachlin2009wearable, plotz2012automatic}, home behavior analysis \cite{wen2016adaptive}, user authentication \cite{derawi2010unobtrusive}, and exercise \cite{direkoglu2012team, sundholm2014smart}. The main tasks of sensor-based HAR are to recognize which activity is being performed, such as sitting, walking, and hammering, given the data provided by the on-body sensors like IMU on the smartphone and smartwatch. A large number of studies have been conducted for sensor-based HAR, which can be categorized into traditional machine learning models and deep learning-based methods. Traditional machine learning models represent feature vectors computed over sub-intervals of sensor data using hand-crafted features in statistical and frequency domains and learn a mapping from feature vectors to activity labels using traditional machine learning classifiers such as random forest and support vector machine (SVM) \cite{bao2004activity,chavarriaga2013opportunity,kwon2018adding}. In the case of deep learning approaches, the raw (or lightly pre-processed) sensor data is fed into a deep neural network, which alone performs the feature extraction and classification. Recently, deep learning methods have achieved state-of-the-art performance in HAR by using convolutional neural networks (CNN) as in \cite{yang2015deep} or by combining them with long short-term memory networks (LSTM) as in \cite{ordonez2016deep}. Even more recently, self-attention approaches have reached state-of-the-art results \cite{mahmud2020human, khaertdinov2021contrastive}.
	
	Although the existing sensor-based HAR methods have achieved impressive results, they have a limitation when handling sensor data from diverse subjects (i.e., users) or applying the well-trained models to unseen users. Many existing studies \cite{cutting1977recognizing, singh2017transforming} have shown that different people perform the same activities in different ways due to their varied personal characteristics and behaviors, which makes user recognition possible, but significant performance degradation of activity recognition can occur. This is observed in practice by the gap in performance when evaluating by leaving out subjects instead of leaving out sessions.
	
	The approaches to this problem can be roughly classified into two categories: classic approaches and deep learning-based methods. Classic approaches include directly selecting user-invariant classical features \cite{saputri2014user} or directly building one model per user \cite{hong2015toward}. These methods can achieve good performance, but they require labeled data to build models for all users, which is not realistic for many HAR applications as it may increase costs and deployment time, and selecting hand-crafted user-invariant features may not be feasible as developing said features requires expertise in the target domain. In addition, this approach may weaken the overall performance depending on the quality of the features available.
	
	Recently, deep learning has been applied to this problem by exploring multi-task or even adversarial learning. In \cite{chen2020metier}, authors exploit user labels by combining HAR with subject identification in a multi-task learning framework that allows the model to focus on the relevant features for each user. This shows that taking user information into account can improve classification results, but it is not clear if models can learn to generalize beyond the available training subjects as they were not evaluated when leaving users out. Still, their work shows that deep learning models can clearly benefit from taking user information into account.  Other works such as \cite{sheng2020weakly} have shown that differences in the environment can, to some extent, be mitigated by employing similarity-based multi-task learning, but they have also not evaluated their model leaving users out and their representation tends to create one cluster per subject with sub-clusters per activity, which may not favor generalization, being more similar to user-models.
	
	In the other direction, \cite{bai2020adversarial} adopted adversarial learning to generate a feature representation more robust regarding user variations. Instead of allowing the model to exploit user-specific information for classification, it should avoid leaking it as there is an adversary (a discriminator) whose objective is to separate subjects in the feature space. They achieved this by employing a Wasserstein Generative Adversarial Network (WGAN) \cite{arjovsky2017WGAN} and Siamese networks and showed that their method could generalize to new subjects without sacrificing performance. This may have other advantages besides performance, as neural networks are known to leak subject information \cite{iwasawa2017privacy} and applying adversarial learning can mitigate those concerns. In addition, other researchers \cite{leite2020improving, soleimani2021cross} improved the performance of the wearable sensor-based human activity recognition by utilizing adversarial learning with the activity classifier network.
	
	While the deep learning-based methods with adversarial learning such as \cite{bai2020adversarial, leite2020improving, soleimani2021cross} have achieved significant improvements, most of these methods still have limitations. Although adversarial learning improved the performance of the activity recognition by generalization, it cannot measure the degree of generalization of the latent features in the training procedure. In the adversarial learning procedure, feature representation modules as generators are trained to fool a discriminator, while the discriminator conducts binary classification to distinguish between two different features from representation modules with randomly chosen two subjects. Thus, it cannot measure the degree of generalization for all subjects and cannot generalize the feature representation over all subjects. 
	In addition, the multi-view data representation module in \cite{bai2020adversarial} comprises three different networks to merge the sub-representations of different views. It requires large amounts of annotated data for human activity recognition to train this complex multi-view data representation module. However, it is challenging to collect a sufficiently large amount of datasets in personalized human activity recognition applications. Furthermore, those works \cite{leite2020improving, soleimani2021cross} can deteriorate the performance  of the activity recognition when the feature extractor or generator is focused on fooling the discriminator.
% 	Annotated medical datasets are limited due to the laborious labeling process, and sometimes legal issues associated with publicly sharing private personal information .	
	
	To overcome these problems, we propose a novel cross-subject adversarial learning for sensor-based human activity recognition. The proposed model is capable of learning a subject-independent embedding feature representation from multiple subjects and generalizing it to unseen target subjects by using an adversarial learning procedure. The adversarial learning between a feature extractor and a discriminator, which distinguishes the extracted features from multiple subjects, learns the distributions of multiple subject data and extracts subject-invariant generalized features for activity recognition. To measure the degree of the feature generalization and align the distributions among the multiple subjects, we use the Maximum Mean Discrepancy (MMD) \cite{gretton2006kernel, li2015generative, long2015learning} regularization. The MMD regularization helps enhance the generalization ability of the proposed adversarial learning method by quantifying the feature generalization. However, the performance of human activity recognition and feature generalization can be different depending on the architecture of the feature extractor network. The previous work \cite{suh2022adversarial} proposed an encoder-decoder structure based on the CNN structure to preserve the characteristic of the original signals, which has been utilized in weakly supervised learning and feature representation \cite{zeng2017semi, varamin2018deep}.
	
	In this paper, we extend the previous work by using transformer network architecture \cite{dosovitskiy2020image, plizzari2021skeleton}, which accounts for individual sensor orientations and spatial and temporal features, for cross-subject human activity recognition. In addition, we adopt the teacher-free self-knowledge distillation \cite{yuan2020revisiting} to improve the stability of the training procedure and balance the optimization between feature generalization and activity recognition. Normally, the self-knowledge distillation trains the same network architecture as the teacher and the student models to regularize the model to avoid overfitting. In the proposed framework, the student model is trained not only with self-knowledge distillation regularization but also with adversarial learning and MMD regularization. In other words, the role of the self-knowledge distillation in the proposed framework is not only to regularize the networks to avoid overfitting to the activity recognition but also to prevent bias to the cross-subject feature generalization by adversarial learning and the MMD regularization.
	Through a series of experiments on four public real-world human activity recognition datasets, we demonstrate the effectiveness of the proposed method. In addition, we evaluate the proposed method a step further that multiple datasets can be combined together. We discuss the overall potential of the proposed model to generalize various domains.
	
	The contributions of this paper can be summarized as follows. 
	\begin{itemize}
		\item A novel Transformer-based Adversarial learning framework for human activity recognition using wearable sensors via Self-KnowledgE Distillation (TASKED) is proposed.
% 		adversarial cross-subject human activity recognition model with a spatial-temporal transformer network is proposed. 
		\item We formulate the adversarial learning between feature extraction and subject discriminator and improve the generalization of the extracted features and the performance of the activity recognition. 
		\item  We use the MMD regularization to enhance the generalization of the feature representation and measure the degree of the generalization.
		\item We adopt the self-knowledge distillation method to improve the stability of the training procedure and balance training optimization between feature generalization and activity recognition.
		\item To validate the proposed method, we conducted experiments with four public benchmark datasets: Opportunity \cite{chavarriaga2013opportunity}, PAMAP2 \cite{reiss2012introducing}, MHEALTH \cite{banos2014mhealthdroid}, and RealDISP \cite{banos2012benchmark}. By the experiments on each and across multiple datasets, we can verify the advantages and effectiveness of the proposed method for feature generalization and activity recognition.
	\end{itemize}
	
	The rest of the paper is organized as follows. Section \ref{sec:relatedworks} introduces the related works.
	Section \ref{sec:proposedmethod} provides the details of the proposed method. Section \ref{sec:experimentalresults} presents quantitative experimental results on the four datasets and their combinations. Finally, Section \ref{sec:conclusion} concludes the paper.
	
	\section{Related works}
	\label{sec:relatedworks}
	
	\subsection{Sensor-based Human Activity Recognition}
	Many researchers have studied sensor-based HAR \cite{dang2020sensor}. The task of sensor-based HAR can be considered as a time-series classification where sensor data is obtained by different types of sensor devices such as inertial measurement units (IMU), electrocardiography (ECG), electromyography (EMG), and heart rate measurement. Bulling et al. \cite{bulling2014tutorial} introduced the general-purpose process framework for activity recognition, which treats individual frames of sensor data as statistically independent. Previous studies can be classified into classical HAR methods and deep learning-based methods. The classical HAR methods have been investigated to extract hand-crafted features to capture the data distributions of activities. The most frequently used hand-crafted features are time-domain features, such as mean, variance, and skewness, and frequency domain features, such as power spectral density \cite{janidarmian2017comprehensive}. Anguita et al. \cite{anguita2012human} proposed a multi-class SVM model on a smartphone to recognize six-class locomotion activities. Hammerla et al. \cite{hammerla2013preserving} introduced an empirical cumulative density function (ECDF) feature to preserve the spatial information of the signal frames. Kwon et al. \cite{kwon2018adding} extended it by adding temporal structures to the ECDF and showed the improvement of activity recognition. 
	
	Recently, deep learning-based HAR methods have been widely explored in existing work \cite{plotz2011feature, lane2015deepear, alsheikh2016deep}, with the fast development and advancement of deep neural networks. Yang et al. \cite{yang2015deep} proposed a human activity recognition method using CNNs, in which the multiple convolutions and pooling filters were designed along the temporal dimensions to process the sensor data. Bhattcharya and Lane \cite{bhattacharya2016sparsification} have introduced a sophisticated model optimization method for constrained resource inference on wearable devices such as smartphones and smartwatches. Contrary to the activity recognition methods using CNNs, recurrent deep learning methods \cite{nakano2017effect} have been researched in the field of HAR. Most prominently used models, so-called LSTMs units \cite{hochreiter1997long}, like other recurrent neural networks, are recurrent neural networks with principally infinite memory for every computing node and have been used very successfully for HAR. Ord{\'o}{\~n}ez and Roggen \cite{ordonez2016deep} proposed a DeepConvLSTM which combines the LSTM model with a number of CNN layers to learn sensor representation by capturing the short-term and long-term temporal correlations for activity recognition. Alsheikh et al. \cite{alsheikh2016deep} proposed a hybrid approach using a deep belief network as an emission matrix of a hidden Markov model to extract features from the sequence of human activities.	In \cite{hammerla2016deep}, appropriate training procedures have been analyzed the basic deep learning approaches including basic temporal CNN and deep LSTM networks through large scale experimentation. Guan and Pl{\"o}tz \cite{guan2017ensembles} proposed Ensembles of deep LSTM networks which combine multiple LSTM networks and achieved better performance than the previous works. More recently, the self-attention mechanism \cite{vaswani2017attention} has been applied for activity recognition. Zeng et al. \cite{zeng2018understanding} applied the self-attention mechanism on the LSTM networks to highlight the important part of time-series signals. Similarly, Mahmud et al. \cite{mahmud2020human} proposed an activity recognition method by employing self-attention, which has reached state-of-the-art results. However, such deep models often require high computation and memory resources. Furthermore, all the works assume that the data among subjects in the training and test datasets follow the same data distributions. In real-world activity recognition applications, different people perform the same activities in different ways, so the data distributions among subjects have a huge gap.
	
	\subsection{Subject-independent Human Activity Recognition}
	
	Generally, there are two ways to capture the interpersonal variability: One is to increase the amount of training data from different subjects, and the another is to extract subject-independent features. The former is too expensive and impossible to collect and annotate data from different people. Recently, transfer learning methods have been investigated to solve cross-domain HAR problems, including cross-sensor-modalities \cite{morales2016deep}, cross-locations \cite{chiang2017feature}, and cross-subjects \cite{handiru2016optimized, zhao2020discriminant}. Domain adaptation is the particular branch of transfer learning, that measures data distribution heterogeneity and aligns among data distributions \cite{cook2013transfer}. Deng et al. \cite{deng2014cross} proposed a cross-person activity recognition method using a reduced kernel extreme learning machine on the source domain, which classifies the target sample and the high confident samples and applies them to the training dataset. Zhao et al. \cite{zhao2011cross} introduced a transfer learning embedded decision tree algorithm that integrates a decision tree and the k-means clustering algorithm to recognize mobile phone-based different personalized activities by model adaptation. Wang et al. \cite{wang2018stratified} proposed a stratified transfer learning method that adopted the pseudo-leveling concept on the unlabeled target data by measuring MMD between the feature spaces for the source domain data and the pseudo-labeled target data. Khan et al. \cite{khan2018scaling} proposed a heterogeneous deep convolutional neural network (HDCNN) that used a feature matching approach to adapt the pre-trained network, trained with the supervised source domain dataset, to an unlabeled target dataset collected from the smartwatch through the minimization of the discrepancy between two datasets after every convolutional and fully connected layer. They used Kullback-Leibler (KL) divergence as a distance measure between the pre-trained network and the target domain feature extractor network. Faridee et al. \cite{faridee2019augtoact} proposed an AugToAct framework which directly aligns data distributions between source and target domains by combining augmentation transformations with deep semi-supervised learning to infer complex activities with the minimal labels in both source and target domains. AugToAct similarly performed domain adaptation as HDCNN, but employed Jensen–Shannon (JS) divergence to minimize the discrepancy among two different domains instead of KL divergence. Akbari and Jafari \cite{akbari2019transferring} extracted stochastic features by training variational auto-encoder instead of deterministic feature extraction and employed the same network architecture of HDCNN to apply the feature matching approach to adapt the model to a target environment. Zhao et al. \cite{zhao2020local} a local domain adaptation method for cross-domain HAR, which aligns the distributions of source and target domains by using the MMD regularization. They first classified the activities into abstract clusters and mapped the original features into a low-dimensional subspace, where the MMD between two clusters with the same label from different domains were minimized. The above domain adaptation approaches consider only a single source domain for the domain adaptation tasks.
	
	Several studies have researched multiple source domain adaptation for sensor-based activity recognition. Some approaches focus on explicitly identifying the most relevant source domain among the multiple source domains with the target domain based on similarity measurements such as cosine similarity. Wang et al. \cite{wang2018deep} proposed a transfer neural network to perform knowledge transfer for activity recognition (TNNAR), which captures both the time and spatial relationship between activities. They considered explicitly selecting the most relevant domain from the multiple source domains based on the cosine similarity and applied the selected domain for domain adaptation to the target domain. Another group of approaches combines all the available source domain data and projects into lower-dimensional space, which is further processed by the classifier. Jeyakumar et al. \cite{jeyakumar2019sensehar} proposed a SenseHAR, which is a sensor fusion model for each device that mapped the raw sensor values to a shared low dimensional latent space. They mitigated the heterogeneous data distribution and assigned labels to the unlabeled data. However, the above-mentioned approaches did not consider multiple source domain data simultaneously, and the approaches cannot capture the uncertainty within the classification tasks.

	Recently, Multi-task and generative adversarial learning (GAN) \cite{goodfellow2014generative}-based methods have been introduced to solve the different data distribution problems. Chen et al. \cite{chen2020metier} proposed a deep multi-task learning-based activity and user recognition (METIER) model, which combines activity recognition and user recognition with a multi-task model. The model shares parameters between the activity recognition module and the user recognition module, and the activity recognition performance can be improved by the user recognition module employing a mutual attention mechanism. Sheng et al. \cite{sheng2020weakly} proposed a weakly supervised multi-task representation learning, which used Siamese networks to exploit a temporal convolutional network as a backbone model. Bai et al. \cite{bai2020adversarial} introduced a discriminative adversarial multi-view network, which extracts multi-view features from temporal, spatial, and Spatio-temporal views using CNN, and generalizes the multi-view features by employing WGAN and Siamese network architecture to decrease the variants between the extracted features from different subjects.
	
	Another general advance in deep neural networks, relevant to this work, is the use of attention and transformers \cite{vaswani2017attention}. Those models have achieved state-of-the-art results in many areas including natural language processing \cite{devlin2018bert}, vision \cite{dosovitskiy2020image}, and skeleton-based HAR \cite{plizzari2021skeleton}. Other works, such as \cite{mahmud2020human} also use self-attention for wearable sensor-based HAR obtaining state-of-the-art results. Still, their model combines data from the different present sensors in the first layers using 1x1 convolutions, forgoing richer explicit attention across sensors further down the network.
	Liu et al. \cite{liu2020giobalfusion} proposed a global attention module for multi-sensor information fusion. The mechanism used the global position attention and modality attention module to selectively boost the influence of informative features and suppress unrelated interference at the fusion layer. Recently, Miao et al. \cite{miao2022towards} proposed a GCN-based dynamic inter-sensor correlations learning framework, named DynamicWHAR, which extracts dynamic features from multi-sensor data by modelling the dynamic correlations between different sensors.
	
	\section{Proposed method}
	\label{sec:proposedmethod}
	    \subsection{Problem Formulation}
	    In this section, we introduce a transformer-based adversarial learning framework for cross-subject sensor-based human activity recognition via self-knowledge distillation. The sensor-based activity recognition task is to use data collected by sensors at different positions on the human body to predict one of $n_a$ activity labels. Let $X = [x_1, ..., x_n]$ be the time-series sensor data obtained by applying a sliding window of size $n_w$ of the $n_c$ available sensor channels  with $x_i \in \mathbb{R}^{n_c \times n_w} $ consisting of the sensor data for that window and $Y = [y_1, ..., y_n]$ the corresponding activity label set of $X$, and $S = [s_1, ..., s_{n}]$ the corresponding subject label set of $X$.
	    We assume our training data $\{X_{src},Y_{src},S_{src}\}$ shares the same $n_a$ activities and sensor types with the test data $\{X_{tgt},Y_{tgt},S_{tgt}\}$ with unseen subjects in the training data used for validation.
    	The goal of the proposed method is to generalize extracted features from the $S_{src}$ source subjects to the target subjects $S_{tgt}$ and improve the performance of the overall activity classification.
    	
	    \subsection{Network Architecture with transformer}
	    \begin{figure*}[!t]
    		\centering
    		\includegraphics[width=\linewidth]{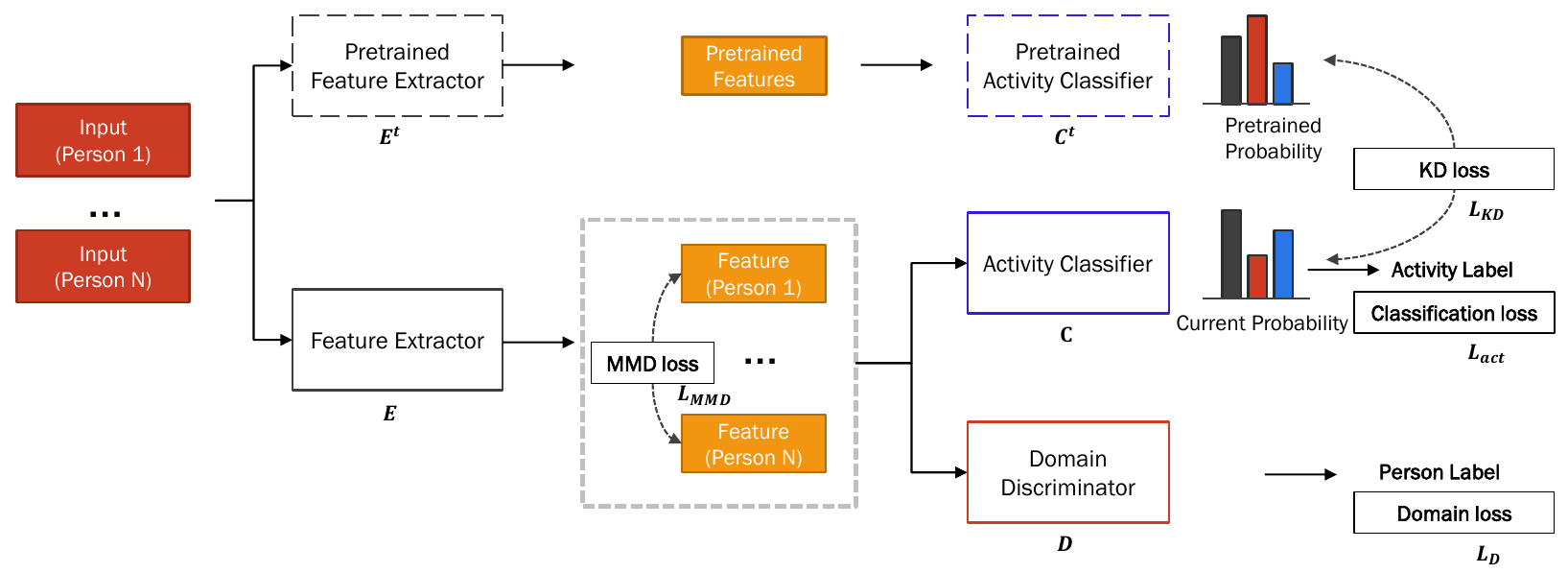}
    		\caption{The overall network architecture of the proposed framework.}
    		\label{fig:overview}
    	\end{figure*}
    	
    	Our goal is to extract cross-subject generalized features to improve the performance of the activity recognition network.
	    The proposed neural network architecture comprises three independent networks: a feature extractor, an activity classifier, and a subject discriminator. The feature extractor represents the proposed transformer network, and the activity classifier maps the features extracted from the feature extractor into the activity labels. The subject discriminator is trained to distinguish the subject label from which the embedding features originated. In contrast, the feature extractor is trained to fool the subject discriminator by providing features that cannot be discriminated between subjects by using an adversarial learning framework. To improve the stability of the training procedure and optimize the balance between the subject discriminator and the activity classifier, we train the proposed overall networks with the self-knowledge distillation method using the pre-trained feature extractor and activity classifier. The overall structure of the proposed framework with the three independent networks is shown in \cref{fig:overview}.
	    
	    % Feature extractor and activity classifier
    	The feature extractor aims to map the input space $\mathcal{X}$ to a common embedding feature space $\mathcal{E}$ and we denote the feature extractor $E: \mathcal{X} \rightarrow \mathcal{E}$. 
    	
    	Our feature extractor works as follows:
    	First, each sensor's data is processed by its independent convolutional layer. This is done to bring all sensors to the same number of channels but has other advantages.
    	Separate convolutional layers are beneficial for HAR \cite{plotz2011feature} and make the architecture more adaptable, as one can add different ones for different sensor modalities or applications.
    	Then, we concatenate all sensor representations, generating a tensor $C \times W \times S$, where $W$ is the size of our sliding window, $S$ is the number of sensors, and $C$ is the selected number of filters for each sensor's convolutional layer. The number of sensors and window size varies per dataset, but we used 32 for $C$ across all of them.
    	
\begin{figure*}[!t]
            \centering
            \includegraphics[width=\textwidth]{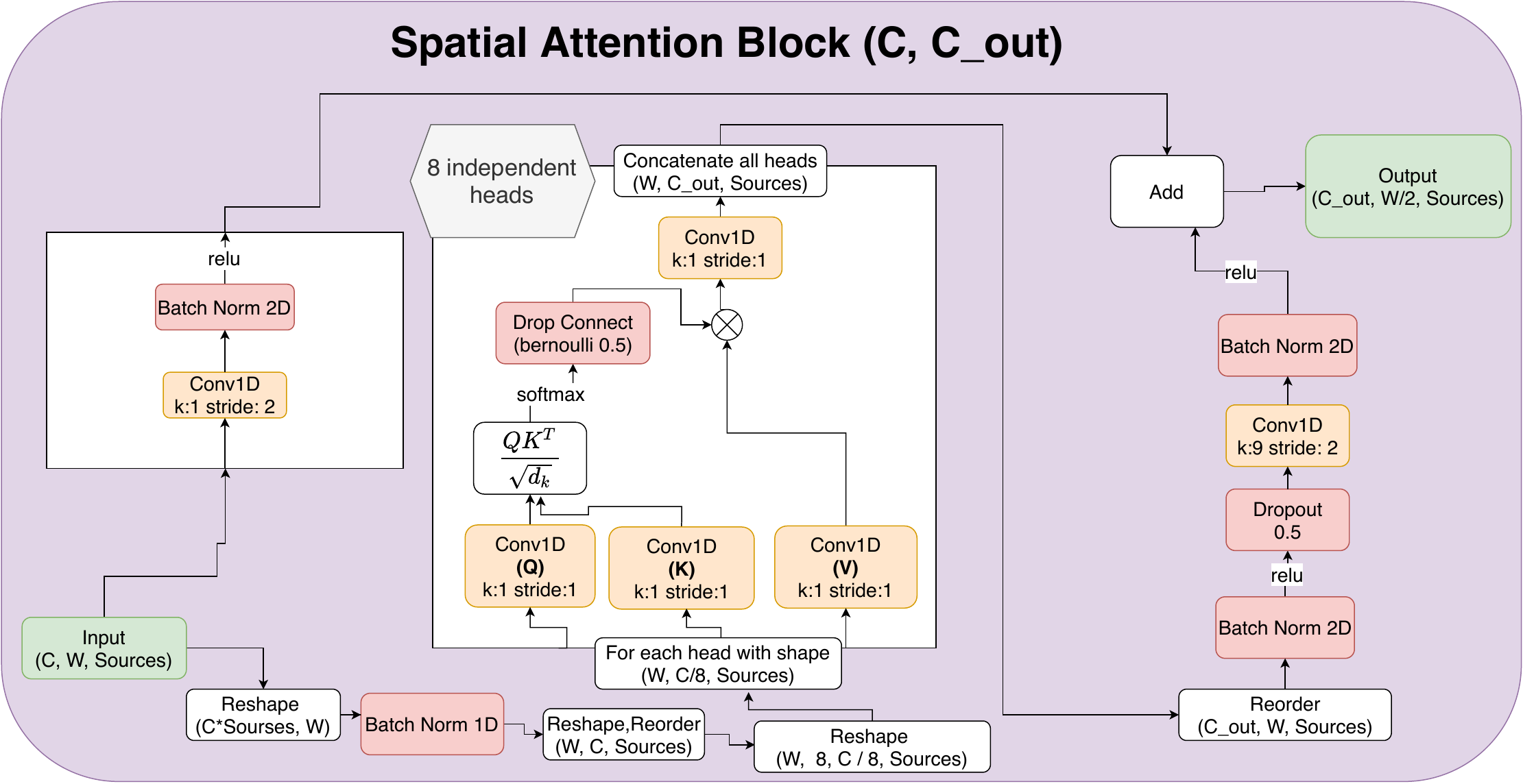}
            \caption{Structure of the attention block used in the transformer feature extractor.}
            \label{fig:block}
        \end{figure*}

    	As transformers have obtained state-of-the-art results in many fields, including skeleton-based HAR \cite{plizzari2021skeleton}, we combine sensor representations using our spatial attention block, which is based on \cite{plizzari2021skeleton}. Our block (\cref{fig:block}) computes attention scores between sensors (spatial attention) and includes temporal convolutions that also reduce the temporal dimension as the number of channels increases. Notice that our Spatial Attention block includes only convolutions with a kernel size of one (network-in-network) for the attention part, while the temporal convolutions have bigger kernels. This is done since spatial attention computes attention weights in the form $S \times S$, that is, across sensors regardless of time. In fact, our preliminary experiments demonstrated that network performance does not improve with bigger kernels for computing query, key, and value. Moreover, in order to avoid overfitting, our blocks use three types of regularization: batch norm, dropout, and drop connect. Unlike dropout, which randomly zeros network outputs, drop connect randomly zeros network weights, in our case weights related to the attention matrix. This is done after the softmax, so it is followed by re-normalization so that weights still sum up to one. We have selected a transformer architecture for our feature extractor given its performance and the activities we want to recognize. In our experiments, we are not interested in information across windows (long-term dependencies). To handle those, one could add an LSTM layer at the end of the feature extractor or opt for a transformer variant that keeps track of such dependencies, such as \cite{wang2019r}.

    	The activity classifier $C$ predicts the activity labels using the embedding features $C: \mathcal{E} \rightarrow \mathcal{Y}$, where $\mathcal{Y}$ denotes the activity label space. In this paper, the activity classifier is designed with an average pooling layer over a time channel and a 1D convolution layer. The loss of the activity classifier influences the feature extractor, as they are both trained to minimize the supervised classification. 
    	
    	% Feature extractor and subject discriminator
    	The subject discriminator distinguishes the subject label from which the embedding feature. The goal of the proposed method is to extract subject-invariant features from multiple domains. In other words, the feature extractor generates a common embedding feature space for multiple subjects. Though the training procedure with the supervised activity classification encourages learning the data distribution of the activities, the extracted embedding features still contain subject-specific information. By adopting an adversarial manner between the subject discriminator and the feature extractor, the subject discriminator is trained to distinguish the subject label from the embedding features and the feature extractor is trained to fool the subject discriminator. A strong subject discriminator can train the feature extractor to generate embedding features that generalize across data from different subjects. The subject discriminator is composed of three discriminator blocks, each including a convolution layer, batch normalization, and a dropout respectively, and two fully connected layers. To align the distributions among the $N+1$ source and target subjects and further generalize the embedding feature representation, we use the MMD \cite{gretton2006kernel, li2015generative, long2015learning} regularization.
    	
    	In addition, the feature extractor and activity classifier are regularized by using self-knowledge distillation from the pre-trained model to improve the stability of the training procedure and balance the optimization between the feature generalization and activity recognition. Following \cite{yuan2020revisiting}, knowledge distillation from the perspective of label smoothing regularization regularizes model training by replacing the one-hot labels with smoothed ones. We deploy the label smoothing regularization following the conceptual ideas of the teacher-free knowledge distillation technique in the training procedure of the proposed framework. Through this regularization, the feature extractor and activity recognition are trained by themselves and are prevented to be biased to the feature generalization. The detailed descriptions of loss functions are addressed in the following subsection.
    	The detailed structure of the proposed adversarial cross-subject networks is shown in \cref{tab:networkarchitecture}.
    	
    	\begin{table*}[!t]
    		\caption{Detailed structure of the proposed TASKED framework. The fourth column indicates the activation function used in the layer while $k$, $W$ and $C$ denote the kernel size, the window size and the number of channels of input sensor signals, respectively.}
    		\label{tab:networkarchitecture}
    		\centering
    		\resizebox{\linewidth}{!}{
    			\begin{tabular}{cccccc}
    				\hline
    				Network & Name & Layers & Act. Func. & Input Tensor & Output Dimension\\
    				\hline
    				\multirow{8}{*}{Feature Extractor} 
    				 & Input for each Sensor & - & - & - & $ C_s \times W $ \\ 
    				 & Conv block (1 per $S$) & per sensor $\times$ {[}Conv $k=3$, st=1{]} & ReLU & Input for each Sensor & $ 32 \times W $ \\
    				 & Concatenation    & -      & -  & The $S$ Conv blocks   & $32 \times W \times S$       \\
                     & Positional Encoding      & positional encoding as proposed in \cite{wang2021translating}.     & -  & Concatenation & $32 \times W \times S$       \\
                     & Spatial Attention 1      & {[Spatial Attention Block $C=32$,$C_{out}=64$]}  & -  & Positional Encoding   & $64 \times W/2 \times S$     \\
                     & Spatial Attention 2      & {[Spatial Attention Block $C=64$,$C_{out}=128$]}      & -  & Spatial Attention 1   & $128 \times W/4 \times S$    \\
                     & Spatial Attention 3      & {[Spatial Attention Block $C=128$,$C_{out}=256$]}      & -  & Spatial Attention 2   & $256 \times W/8 \times S$    \\
                     & Average Pooling over $S$ & -      &    & Spatial Attention 3   & $256 \times W/8$     \\ 
    				\hline			
    				\multirow{2}{*}{Activity Classifier}
    				& Average Pooling over $W$		& -	 & -		& Average Pooling over $S$ & $1\times$ 256\\
    				& FC 1 		  	& Fully Connected 					   & - 		    & FC 1			& $n_a$ \\
    				\hline
    				\multirow{5}{*}{Subject Discriminator}
    				& Discriminator Block 1 & Conv $k=5$, st=2		& Leaky ReLU & Average Pooling over $S$ 	& 32 $\times~W/16$\\
    				& Discriminator Block 2 & Conv $k=5$, st=2		& Leaky ReLU & Discriminator Block 1 	& 64 $\times~W/32$\\
    				& Discriminator Block 3 & Conv $k=5$, st=2		& Leaky ReLU & Discriminator Block 2 	& 128 $\times~W/64$\\
    				& FC 1 					& Fully Connected					   & ReLU		& Discriminator Block 3	& 10 \\
    				& FC 2 		  			& Fully Connected 					   & - 		    & FC 1			& $N + 1$ \\	
    				\hline		
    		\end{tabular}}
    	\end{table*}

	    \subsection{Loss functions}
	    % Loss functions: MMD loss, classification loss with KD, subject discriminator loss
    	As shown in \cref{fig:overview}, we define three loss functions to train the three independent networks: 1) a classification loss $\mathbb{L}_{cls}$, 2) a domain loss $\mathbb{L}_D$, and 3) an MMD loss $\mathbb{L}_{MMD}$. Let $E$, $C$, and $D$ denote the feature extractor, the activity classifier, and the subject discriminator, respectively. 
    	
    	In the proposed network architecture, the feature extractor and activity classifier are basically trained by supervised learning with given activity labels. Normally, the data distribution of classes in the human activity recognition datasets is often imbalanced. Training a network under imbalanced data conditions can produce a detrimental effect on human activity recognition performance and a biased classification result. To improve the performance of the performance under imbalanced data conditions, we adopt a weighted cross-entropy loss and a dice loss function for activity labels as the initial classification loss.
    	The initial classification loss for the feature extractor and activity classifier is expressed as follows:
    	
    	\begin{equation}
    		\label{eq:classification}
    		\begin{split}
    		    \mathbb{L}_{act} (X,Y) &=  \frac{1}{2} (- \sum_{i=1}^{n_a} w_{y_i} y_i \log C(E(x_i))) \\
    		    &+ \frac{1}{2} (1 - \frac{2\sum\sum y_i \odot \tilde{y}_i + \epsilon}{\sum y_i + \sum \tilde{y}_i + \epsilon}),
    		\end{split}
    	\end{equation}
    	where the first term is the weighted cross-entropy classification loss, the second term is the dice classification loss, $w_i$ is a class weight for each class label, $y_i$ is a one-hot vector of the ground truth of the activity label, $\tilde{y}_i$ is an output one-hot vector from the activity classifier, and $\odot$ denotes an element-wise multiplication. The weighted cross-entropy and dice loss are usually used to solve the data imbalance problem. $w_i$ is calculated with the number of each class instance and the maximum number of data samples among the class distributions $w_i = \frac{\max{(M)}}{m_i}$ where the distribution of the number of data samples per class for the training data is given $M=[m_1, ... , m_{n_a}]$.
    	
    	The knowledge distillation technique transfers knowledge from a teacher model to a student model so that the performance of the student model is improved. The student model learns from more informative sources -the predictive probabilities of the teacher model - instead of just one-hot labels. Following \cite{yuan2020revisiting}, we first train the student model containing the feature extractor and activity classifier in the normal way to obtain a pre-trained model by minimizing \cref{eq:classification}. Hinton et al. \cite{hinton2015distilling} proposed to use temperature scaling to soften the predictive probabilities for better distillation. Given the pre-trained models $E^t$ and $C^t$, the output prediction of the pre-trained network is expressed as follows.
    	\begin{equation}
    	    \label{eq:temperature}
    	    p^t_i(x;\tau) = softmax(z^t_i(x;\tau)) = \frac{\exp{(z^t_i(x)/\tau)}}{\sum_j \exp{(z^t_j(x)/\tau)}}
    	\end{equation}
    	where $z^t$ denotes the output logit vector of the pre-trained model and $\tau$ denotes a temperature parameter. The idea of self-knowledge distillation is to regularize the pre-trained model and the student model by using the Kullback-Leibler (KL) divergence. Namely, a knowledge distillation (KD) regularization loss is defined as follows.
    	\begin{equation}
    	    \label{eq:regularizationloss}
    	    \mathbb{L}_{KD} (X;\theta_E, \theta_C, \tau) = {KL} (p^t(x;\tau), p(x;\tau))
    	\end{equation}
    	where $KL$ is the KL divergence, and $p^t$, $p$ are the output probability of the pre-trained models $C^t(E^t(x))$ and the student models $C(E(x))$, respectively.
    	Similar to the original knowledge distillation method, we try to minimize the initial classification loss and the KD regularization loss between the predictions of pre-trained and student model.
    	\begin{equation}
    	    \label{eq:finalclassification}
    	    \mathbb{L}_{cls} = (1-\alpha) \mathbb{L}_{act} + \alpha \mathbb{L}_{KD}
    	\end{equation}
    	where $\alpha$ is a hyperparameter.
    	
    	Moreover, the goal of the subject discriminator is to distinguish which subject generated the embedding feature representations. The subject discriminator is thus trained to minimize the domain loss:
    	\begin{equation}
    		\label{eq:domain}
    		\begin{split}
    			\mathbb{L}_D (X,S) =  -\sum_{i=1}^{N} s_i \log D(E(x_i)),
    		\end{split}
    	\end{equation}
    	where $s_i$ is the ground truth of the subject label and the equation is the cross-entropy classification loss.
    	
    	Here, in addition to adversarial learning, we address the MMD regularization to align the distributions among different subjects and to further improve the generalization of embedding features extracted by the feature extractor. The MMD is one of the most commonly used non-parametric methods to measure the distance of the distribution between two different domain datasets. The feature extractor represents the embedding features from the input signals $E: \mathcal{X} \rightarrow \mathcal{E}$, and we let $E(X_s) = E_s=[e_{s,1}, e_{s,2},...,e_{s,M}]$ and $E(X_t) = E_t=[e_{t,1},e_{t,2},...,e_{t,N}]$ represent the embedding representations of two different subject domains. A mapping operation $\phi(\cdot)$ projects the representations of two different domains onto the reproducing kernel Hilbert space (RKHS) $\mathcal{H}$ \cite{gretton2006kernel}, and calculates the mean distance between the two domains in RKHS. The MMD between two subject domains can be calculated by using the function $\phi(\cdot)$ as follows:
    	
    	\begin{equation}
    		\label{eq:MMD2}
    		\begin{split}
    			MMD(E_s, E_t)^2 = \left\lVert \frac{1}{M} \sum_{i=1}^{M} \phi(e_{s,i}) - \frac{1}{N} \sum_{j=1}^{N} \phi(e_{t,j}) \right\rVert_{\mathcal{H}}^2
    		\end{split}
    	\end{equation}
    	The key to calculate the MMD is to find the appropriate $\phi(\cdot)$ as a mapping function to map the two domains to RKHS $\mathcal{H}$. Thus, the mean difference between the two data distributions after the mapping is calculated as their difference. $\phi(\cdot)$ represents a characteristic kernel function as $k(e_{s,i}, e_{t,j}) = \langle \phi(e_{s,i}), \phi(e_{t,j})\rangle$, and the MMD can be rewritten as follows:
    	
    	\begin{equation}
    		\label{eq:MMD3}
    		\begin{split}
    			&MMD(E_s, E_t)^2 = \frac{1}{M} \sum_{i=1}^{M} \sum_{j}^{M} k(e_{s,i},e_{s,i}) \\
    			&+ \frac{1}{N} \sum_{i=1}^{N} \sum_{j}^{N} k(e_{t,i},e_{t,i}) - \frac{2}{MN} \sum_{i=1}^{M} \sum_{j=1}^{N} k(e_{s,i}, e_{t,j})
    		\end{split}
    	\end{equation}
    	
    	Generally, the Gaussian kernel function $k(e_{s,i}, e_{t,j}) = \exp(-\frac{\sigma}{\lVert e_{s,i} - e_{t,j} \rVert^2 })$ is used as the kernel function in the MMD algorithm, which maps data to infinite-dimensional space. This MMD method is based on a single kernel transformation. In this work, we adopt the multi-kernel MMD (MK-MMD) \cite{long2015learning}, which is an extension of the MMD and the optimal kernel can be obtained by linear combination of multiple kernels. The kernel function is defined as the convex combination of $m$ positive semi-definite (PSD) kernels ${k_u}$. The total kernel $k$ is defined as follows.
    	
    	\begin{equation}
    		\label{eq:MMDkernel}
    		\begin{split}
    			K \triangleq \Big\{ k=\sum_{u=1}{m} \beta_u k_u : \sum_{u=1}^{m} \beta_u = 1, \beta_u \geq 0, \forall u \Big\}
    		\end{split}
    	\end{equation}
    	where $k$ is weighted by different kernel $k_u$, $\{ \beta_u\}$ is the coefficient, which is the weight of $K$, to ensure that the generated multi-kernel $k$ is characteristic. 
    	
    	Unlike the general domain adaptation, the goal of the proposed method is to generalize the features from multiple subject domains from either only the $N$ source subjects or both $N+1$ source and target subjects. Thus, the overall MMD regularization loss $\mathbb{L}_{MMD}$ is described as follows.
    	
    	\begin{equation}
    		\label{eq:overallMMD}
    		\begin{split}
    			\mathbb{L}_{MMD} (E_1,...,E_{N+1})	=  \frac{1}{(N+1)^2} \sum_{1 \leq i,j \leq N+1} MMD(E_i,E_j)
    		\end{split}
    	\end{equation}
    	
    	% Adversarial loss between feature extractor and subject discriminator
    	The goal of the feature extractor is to extract the generalized feature among different subject domains, preserve the characteristics of the original data, and learn class-discriminative feature representations. In other words, the feature extractor is trained to jointly minimize the losses of classification and MMD, and maximize the domain loss for the adversarial learning, simultaneously, whereas the activity classifier and the subject discriminator are trained to minimize the classification loss and domain loss, respectively. Finally, The objective functions of the proposed method are defined as follows:
    	
		\begin{equation}
			\label{eq:objective}
			\begin{split}
				\min_{E,C} &\max_{D} \mathbb{L}_{obj} (X,Y,S) = \lambda_{cls} \mathbb{L}_{cls} (X,Y) + \lambda_{MMD} \mathbb{L}_{MMD} (X)\\
				 &- \lambda_{D} \mathbb{L}_D (X,S)
			\end{split}
		\end{equation}
    % 	\begin{equation}
    % 		\label{eq:objective}
    % 		\min_{E,C} \max_{D} \mathbb{L}_{obj} (X,Y,S)
    % 	\end{equation}
    % 	where $\mathbb{L}_{obj}$ is
    % 	\begin{equation}
    % 		\lambda_{cls} \mathbb{L}_{cls} (X,Y) + \lambda_{MMD} \mathbb{L}_{MMD} (X) - \lambda_{D} \mathbb{L}_D (X,S)
    % 	\end{equation}
    	where $\lambda_{cls}$ controls the relative importance of activity classification, and $\lambda_{MMD}$ and $\lambda_{D}$ are hyperparameters that control the effect of domain generalization.     
    	
    	In summary, the classification loss $\mathbb{L}_{cls}$ is used to improve the performance of the activity recognition with the regularization technique of the self-knowledge distillation, and the MMD regularization term $\mathbb{L}_{MMD}$ helps to measure and align the distribution distance among different subjects, and the domain loss $\mathbb{L}_{D}$ hinders extracting subject domain-specific information from the feature extractor by the adversarial learning between the feature extractor and subject discriminator.
	    
	    \subsection{Training Procedure}
	    \label{subsec:training}
    	% Figure and Algorithm
    	\begin{figure*}[!t]
    		\centering
    		\includegraphics[width=\linewidth]{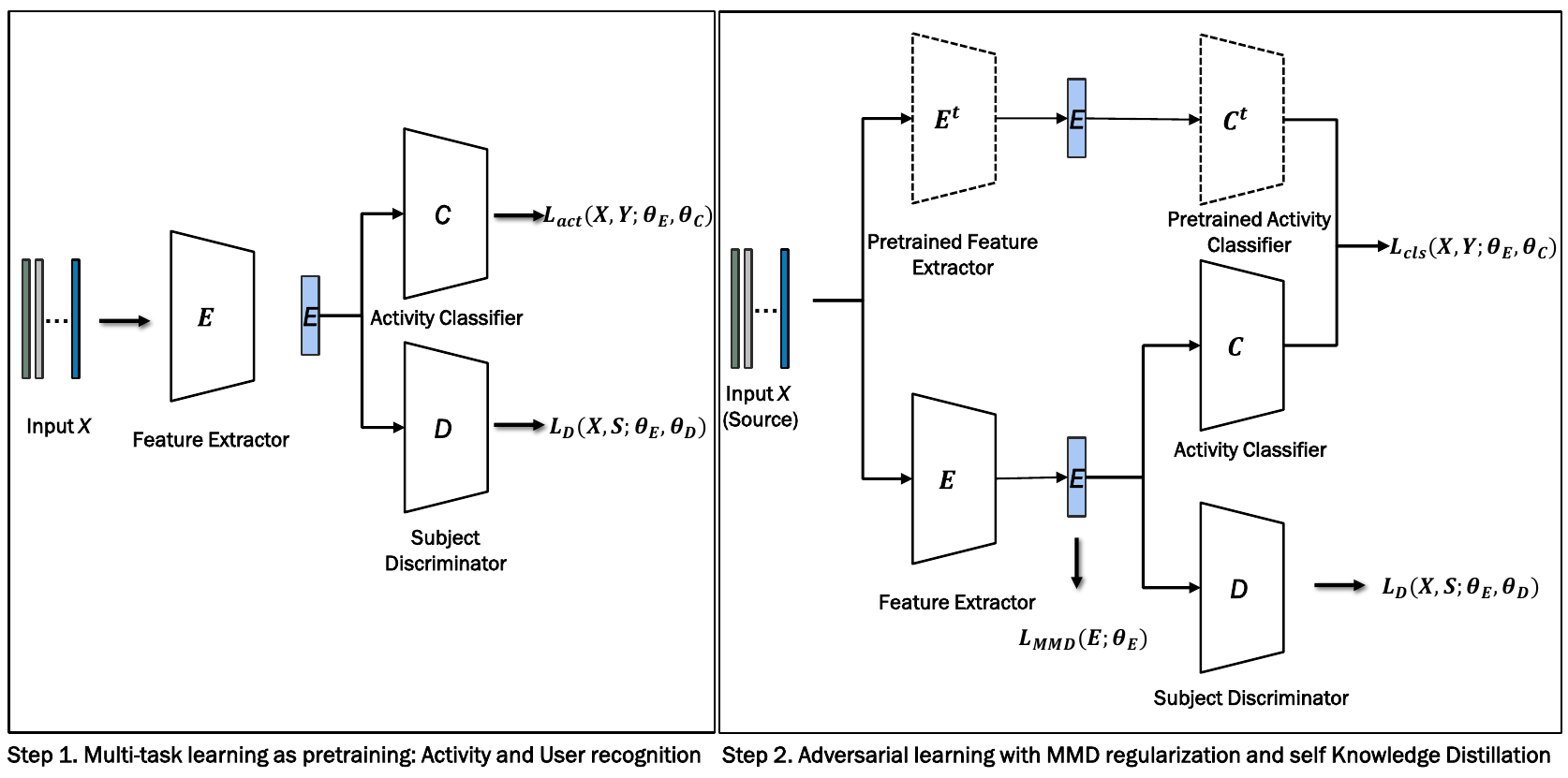}
    		\caption{Training procedure, representing the two steps described in \cref{subsec:training}. Solid lines indicate that the network is being trained, and dashed lines indicate that the parameters of the network are fixed.}
    		\label{fig:procedure}
    	\end{figure*}
    	
    	\begin{algorithm}[!h]
    		\caption{Training procedure for TASKED framework. We use default values of $\lambda_{cls}=10$, $\lambda_{MMD}=5$, $\lambda_D=1$, $\alpha=0.6$, $\tau=20$}\label{Algo}
    		\begin{algorithmic}[1]
    			\REQUIRE Batch size $m$, Adam hyperparameter $\eta$, hyperparameters $\lambda_{cls}$, $\lambda_{D}$, $\lambda_{MMD}$.
    			\STATE \textbf{Input:} $X$, $Y$, $S$, $X_t$, $Y_t$, $S_t$
    			\FOR{number of training iteration for step 1}
    			\STATE Sample a batch $(x,y,s)$ from the training dataset $X$, corresponding activity label $Y$, and domain label $S$.
    			\STATE $\theta_C \gets \theta_C - \eta_C \nabla_{\theta_C} \mathop{\mathbb{L}_{act}}(x,y;\theta_C)$ \hfill$\triangleright$\cref{eq:classification} 
    			\STATE $\theta_D \gets \theta_D - \eta_D \nabla_{\theta_D} \mathop{\mathbb{L}_D}(x,s;\theta_D)$ \hfill$\triangleright$\cref{eq:domain}
    			\STATE $\theta_E \gets \theta_E - \eta_E \nabla_{\theta_E} (\mathop{\mathbb{L}_{D}} + \mathop{\mathbb{L}_{act}(x,y,s;\theta_E)})$ \hfill $\triangleright$\cref{eq:classification,eq:domain}
    			\ENDFOR
    			\FOR{number of training iteration for step 2}
    			\STATE Sample a batch $(x,y,s)$ from the training dataset $X$, corresponding activity label $Y$, and domain label $S$.
    			\STATE $\theta_D \gets \theta_D - \eta_D \nabla_{\theta_D} \mathop{\mathbb{L}_D}(x,s;\theta_D)$ \hfill$\triangleright$\cref{eq:domain}
    			\STATE $\theta_C \gets \theta_C - \eta_C \nabla_{\theta_C} \mathop{\mathbb{L}_{cls}}(x,y;\theta_C)$ \hfill$\triangleright$\cref{eq:finalclassification}
    			\STATE $\theta_E \gets \theta_E - \eta_E \nabla_{\theta_E} \mathop{\mathbb{L}_{obj}}(x,y,s;\theta_E)$ \hfill $\triangleright$\cref{eq:objective}
    			\STATE Sample a batch $(x_t,s_t)$ from the target dataset $X_t$, and corresponding domain label $S_t$.
    			\STATE $(x', s')  \gets (concat(x, x_t), concat(s, s_t))$
        		\STATE $\theta_D \gets \theta_D - \eta_D \nabla_{\theta_D} \mathop{\mathbb{L}_D}(x',s';\theta_D)$ \hfill $\triangleright$\cref{eq:domain}
    			\STATE $\theta_E \gets \theta_E - \eta_E \nabla_{\theta_E} \mathop{(\mathbb{L}_{MMD}} - \mathop{\mathbb{L}_D}(x',s';\theta_E))$ \hfill $\triangleright$\cref{eq:domain,eq:overallMMD}
    			\ENDFOR
    		\end{algorithmic}
    	\end{algorithm}
	
    	The training procedure for the proposed TASKED framework is based on two steps in order to train the three independent neural networks stably. \cref{fig:procedure} presents the training procedure with the detailed steps. The first step is multi-task learning as a pre-training step for activity and subject recognitions. In this step, we jointly minimize the losses of activity classification and subject discrimination to train the feature extractor to capture the data distribution of each activity and the data distributions among subjects. At the same time, the subject discriminator is trained to minimize the discriminator loss \cref{eq:domain} and the activity classifier is trained to minimize the classification loss \cref{eq:classification}. Finally, we train the adversarial learning framework via MMD regularization and self-knowledge distillation. In this step, we compute the final classification loss \cref{eq:finalclassification} using the KD regularization between the pre-trained and the student model. While the feature extractor is trained to minimize the subject discriminator loss in the first step, the feature extractor is trained to maximize the discriminator loss in the second step for the adversarial learning scheme. The training details for the proposed adversarial feature extraction method are summarized in \cref{Algo}.
	
	\section{Experimental results}
	\label{sec:experimentalresults}
	\subsection{Datasets and Evaluation Metrics}
		To evaluate the effectiveness of the proposed method for human activity recognition, four types of popular HAR public datasets were used, Opportunity \cite{chavarriaga2013opportunity}, PAMAP2 \cite{reiss2012introducing}, MHEALTH \cite{banos2014mhealthdroid} and RealDISP \cite{banos2012benchmark}, that contain continuous sensor data of various sensors and different human activities by different participants.
    	
        \begin{itemize}
    		\item Opportunity: The Opportunity dataset is for human activity recognition from wearable, object, and ambient sensors. It contains annotated recordings from four users with 7 inertial measurement units (IMU) and 12 on-body accelerometers. The dataset was recorded with 6 runs per user, that 5 runs are the activity of daily living runs (ADL)  by a natural execution of daily activities and the last run is called a drill run by a scripted sequence of activities. The dataset contains a total of 6 hours of recordings. The provided sampling frequency of all IMU sensors is $\SI{30}{\hertz}$. The dataset comprises three types of sensors: body-worn sensors with 145 channels, object sensors with 60 channels, and ambient sensors with 37 channels. In this paper, we selected a dimension of 113 channels taking into account only the body-worn sensors including the IMUs and accelerometers, following the setup of \cite{ordonez2016deep}. We preprocessed all channels of sensor data to fill in missing values using linear interpolation and to normalize the data values per channel to interval $[0, 1]$ with manually set minimum and maximum values per channel as in \cite{ordonez2016deep}. We used a sliding window size of 64 with a sliding step of 16, which is close to two seconds of the sliding window and a 0.5-second step size. We use two types of annotations from the dataset. One is modes of locomotion and postures, such as Stand, Walk, Sit, and Lie is annotated with five classes. Another is 18 mid-level gestures such as Open Door, Close Door, and Clean Table. 
    		
    		\item PAMAP2: The PAMAP2 dataset was recorded from nine participants that were instructed to perform 18 activities of daily living. The dataset contains a total of more than 10 hours of recordings. One channel of heart rate-monitor and three IMUs were placed on the subject's chest, dominant wrist, and dominant ankle. The dataset has 52 channels, containing a channel of heart rate, 17 channels per IMU. The full IMU sensory data is composed of 6 channels of acceleration data, 3 channels of gyroscope data, 3 channels of magnetometer data, and 3 channels of orientation. In this work, we selected a total dimension of 36 channels by removing a channel of heart rate, a channel of temperature per IMU, 4 channels of orientation per IMU, since the orientation of IMUs is mentioned as invalid in the data collection. Additionally, we remove six activities classified "Optional" in the dataset and the ninth subject, since the "Optional" activities were collected by only one subject and the ninth subject executed only one activity. Thus, a total of 12 activities named "protocol" in the dataset from 8 subjects are used in this work. We preprocessed all channels of selected sensor data to fill in NaN values using linear interpolation. All channels were normalized to zero mean and unit variance per user. The IMU data were collected under the sampling frequency of $\SI{100}{\hertz}$ and we used a sliding window length of 200 (2 seconds) with a sliding step of 50 (0.5 second).
    		
    		\item MHEALTH: The MHEALTH dataset contains basic body movements and vital signs data recorded from 10 volunteers of diverse profiles. The volunteers carried out 12 physical activities such as climbing stairs, walking, waist bends forward, and cycling. Each activity was recorded for a minute or repeated 20 times. Three IMUs were placed on the subject's chest, right wrist, and left ankle to measure acceleration, rate of turn, and magnetic field orientation. Additionally, the sensor was positioned on the chest to provide 2-lead ECG measurements, which can be used for basic heart monitoring or checking the effects of exercise on the ECG. The provided sampling rate of all sensing modalities is $\SI{50}{\hertz}$. To evaluate the proposed method with the iterative leave-one-subject-out cross-validation procedure, we augmented the dataset with a sliding window length of 200 (4 seconds) and a step size of 50 (1 second), unlike other methods \cite{nguyen2015recognizing, sheng2020weakly} used a sliding window length of 5 seconds and a step size of 2.5 seconds.
    		
    		\item RealDISP: The RealDISP dataset consists of a total of 9 hours of daily activity data from 17 volunteers. The volunteers were instructed to perform 33 activities involving the whole-body movements and body part-specific activities such as walking, jogging, jumping, cycling, knee bending, waist bending, and rowing. 9 IMUs were distributed on the arms, wrists, calves, thighs, and back of the subject. Each sensor provides 3 channels of acceleration, 3 channels of gyroscope, 3 channels of magnetic field measurements, and four channels of quaternions for orientation. The sensor data were collected under the sampling rate of $\SI{50}{\hertz}$ and we used a sliding window length of 120 (2.4 seconds) with a sliding step of 60 (1.2 seconds).
    	\end{itemize}
    	
    	\begin{table*}[!t]
    		\caption{Evaluation dataset information for our experiments}
    		\label{tab:dataset}
    		\centering
    		\resizebox{\linewidth}{!}{
    		\begin{tabular}{lcccccc}
    			\hline
    			Dataset & Subjects & Activities & Channels & Frequency & Window size & Sliding step \\ 
    			\hline
    			Opportunity - locomotion & 4 & 5 & 113 & $\SI{30}{\hertz}$ & 64 & 16 \\
    			Opportunity - gestures & 4 & 18 & 113 & $\SI{30}{\hertz}$ & 64 & 16 \\
    			PAMAP2 & 8 & 12 & 36 & $\SI{100}{\hertz}$ & 200 & 50 \\
    			MHEALTH & 10 & 12 & 23 & $\SI{50}{\hertz}$ & 200 & 50 \\
    			RealDISP & 17 & 33 & 81 & $\SI{50}{\hertz}$ & 120 & 60 \\
    			\hline
    		\end{tabular}}
    	\end{table*}
    	
    	The detailed information of the four datasets is summarized in \cref{tab:dataset}. To evaluate the performance of the proposed model and how much performance varies depending on the subject, we conduct two different experiments: 1) leave-one-subject-out cross-validation procedure on a single dataset, 2) leave-one-subject-out cross-validation procedure across different datasets. Unlike the experimental settings in other methods \cite{ordonez2016deep, chen2020metier}, the leave-one-subject-out cross-validation procedure is that all data from a subject are used as a test set and all data from another subject are used as a validation set, while all data from other subjects are used as a training set. We picked up all data from the first two subjects (that are not the test one) as validation set. Thus, there are two iterations for a test set, but with two different validation sets. For instance, if there are 10 subjects in the given dataset, the training and validation procedure would be conducted 20 times.
    	The second experiment is the leave-one-subject-out across different datasets. The leave-one-subject-out scheme across different datasets is that one subject data from each dataset is used as a test set and another subject data from each dataset is used as a validation dataset while the rest data from each dataset are used as a training set. We selected common activity classes, sensor positions, and sensor data types among different datasets. The subject labels are given from each dataset. For example, the Opportunity dataset has 4 subjects, PAMAP2 8 subjects, and MHEALTH 10 datasets. We set the subject labels of the opportunity to the global labels from 0 to 3, the subject labels of PAMAP2 to the global labels from 4 to 11, and the subject labels of MHEALTH to the global labels from 12 to 21. The evaluation was repeated two times on each test set.

    	To evaluate and compare the performance of the proposed method with others, we adopted three evaluation metrics, which are used in various human activity recognition studies\cite{ordonez2016deep, bai2020adversarial}: accuracy $acc$, weighted F1-score $F_w$, and macro F1-score $F_m$.
    	
    	\begin{equation}
    		\label{eq:accuracy}
    		acc = \frac{TP}{TP + FN + FP + TN} 
    	\end{equation}
    	\begin{equation}
    		\label{eq:weightedF1}
    		F_w = \sum_{c=1}^{C} w_i \frac{2\times precision_c \times recall_c}{precision_c + recall_c}
    	\end{equation}
    	\begin{equation}
    		\label{eq:MAPE}
    		F_m = \frac{2}{C} \times \sum_{c=1}^{C} \frac{precision_c \times recall_c}{precision_c + recall_c}
    	\end{equation}
    	where $Recall=\frac{TP}{TP+FN}$, $Precision=\frac{TP}{TP+FP}$, and $TP$, $FP$, $TN$, and $FN$ denote the true positive, false positive, true negative ,and false negative values,	$C$ denotes the number of classes and $w_i=n_i/N$ is the proportion of samples of the $i$-th class with $n$ being the number of samples of the $i$-th class and $N$ being the total number of samples. 
    	
	\subsection{Implementation Details}
	    The experiments were all implemented using Python scripts in the PyTorch framework. Training procedures were conducted in the Linux system with four NVIDIA Quadro RTX 8000 GPUs. The hyperparameters for \cref{eq:overallMMD} were $\lambda_{cls}=10$, $\lambda_{MMD}=5$, and $\lambda_{D}=1$. In addition, the hyperparameters for \cref{eq:temperature} and \cref{eq:finalclassification} were $\tau=20$ and $\alpha=0.6$. Through various testing, it is observed that the parameters mentioned in the paper yield the best performance. We chose the Adam optimizer \cite{kingma2014adam} with a learning rate of $\eta_C=1 \times 10^{-4}$, $\eta_E=1 \times 10^{-4}$, $\eta_D= 1 \times 10^{-3}$, $\beta_1=0.9$, and $\beta_2=0.99$. The batch size was 128 and the training epochs are 200. Early stopping was used with the patience of 20 epochs to avoid overfitting. For the MK-MMD, the Gaussian kernel is applied to the MK-MMD, and its number is set to 5. 
	
	\subsection{Comparison Results on a Single Dataset}
    	The proposed TASKED framework was evaluated on the Opportunity, PAMAP2, MHEALTH, and RealDISP datasets. The three evaluation metrics were used to evaluate and compare the proposed method to deep-learning-based state-of-the-art methods including multi-channel time-series convolutional neural networks (MC-CNN) \cite{yang2015deep}, DeepConvLSTM \cite{ordonez2016deep}, Self-attention activity recognition method \cite{mahmud2020human}, METIER model \cite{chen2020metier}, and the previous method "Adversarial CNN" \cite{suh2022adversarial}. MC-CNN is a CNN-based model consisting of three convolutional layers, two pooling layers, and two fully connected layers. DeepConvLSTM is a combined model of CNN and LSTM for activity recognition, that comprises four convolutional layers and two LSTM layers to learn both spatial and temporal correlations. Self-attention activity recognition method introduced a self-attention mechanism to improve the performance of the human activity recognition based on wearable sensors. The METIER model is to solve activity recognition and user recognition tasks jointly and transfer knowledge across them by sharing parameters between activity recognition and user recognition networks softly and employing mutual attention mechanism to exploit their knowledge to highlight important features for each other. Lastly, the previous method, Adversarial CNN, proposed an adversarial feature extraction for user independent human activity recognition based on CNN architecture. 
    % 	In addition, to show the effectiveness of the proposed adversarial learning with MMD regularization and self-knowledge distillation, we evaluate the proposed transformer network alone. This serves as an ablation study, highlighting how much of the improvement comes from the new transformer architecture as opposed to our proposed adversarial framework. 
    	For a fair comparison study, we set up the same experimental conditions, such as splitting the training and test datasets, data preprocessing, the window size, and step size, as we addressed in the previous section.
    	
    	The Opportunity dataset is normally evaluated using two of its label types: the modes of locomotion recognition task and the gesture recognition task. We conducted the comparison experiment and introduce the experimental results on the two tasks separately.

        \begin{table*}[!t]
    		\caption{Comparison results with the state-of-the-arts on Opportunity, PAMAP2, MHEALTH, and RealDISP datasets. The numbers are expressed in percent and represented as $mean \pm std$.}
    		\label{tab:comparisonresults}
    		\centering
    		\renewcommand{\arraystretch}{0.8}
    		\resizebox{\linewidth}{!}{
    		\begin{tabular}{lllccc}
    			\hline
    			Dataset & Task	&Method 			& $acc$				& $F_w$				& $F_m$\\
    			\hline
    			\multirow{12}{*}{Opportunity} &	
    			\multirow{6}{*}{Locomotion} 	
    			& MC-CNN 			& 65.79 $\pm$ 8.73	& 63.59 $\pm$ 9.54	& 62.32 $\pm$ 11.00\\
    			& &DeepConvLSTM	    & 66.02 $\pm$ 5.23	& 65.39 $\pm$ 5.08	& 62.88 $\pm$ 10.47\\
    			& &Self-Attention   & 61.71 $\pm$ 21.66	& 60.17 $\pm$ 22.21	& 55.27 $\pm$ 20.32\\
    			& & METIER          & 73.97 $\pm$ 3.84  & 73.62 $\pm$ 3.84  & 74.24 $\pm$ 3.98 \\
    			& & Adversarial CNN & 73.11 $\pm$ 2.98  & 73.06 $\pm$ 3.04  & 72.78 $\pm$ 7.73 \\
     			& & TASKED		& \textbf{75.83 $\pm$ 2.54} & \textbf{75.83 $\pm$ 2.57}	& \textbf{77.09 $\pm$ 3.41}\\
    			\cline{2-6}
    			&\multirow{6}{*}{Gestures} 
    			&MC-CNN			    & 69.47 $\pm$ 5.96	& 68.20 $\pm$ 4.79	& 27.86 $\pm$ 7.39\\
    			& &DeepConvLSTM	    & 71.37 $\pm$ 4.23	& 71.23 $\pm$ 3.37	& 35.59 $\pm$ 7.36\\
    			& &Self-Attention   & 74.09 $\pm$ 5.30	& 69.64 $\pm$ 5.29	& 28.89 $\pm$ 12.79\\
    			& & METIER          & 71.49 $\pm$ 4.94  & 73.31 $\pm$ 3.45  & 44.44 $\pm$ 6.27 \\
    			& & Adversarial CNN & 76.25 $\pm$ 2.42  & 75.54 $\pm$ 2.13  & 44.65 $\pm$ 4.54 \\
    			& &TASKED		& \textbf{80.20 $\pm$ 3.55} & \textbf{79.75 $\pm$ 3.40}	& \textbf{54.40 $\pm$ 6.66}\\
    			\hline
    			\multirow{6}{*}{PAMAP2} 
    			& &  MC-CNN			& 76.50 $\pm$ 15.77	& 75.12 $\pm$ 17.48 & 67.84 $\pm$ 16.28\\
    			& & DeepConvLSTM	& 66.43 $\pm$ 18.74	& 64.69 $\pm$ 20.19	& 57.06 $\pm$ 17.03\\
    			& & Self-Attention  & 76.92 $\pm$ 15.45 & 76.01 $\pm$ 16.58 & 68.68 $\pm$ 15.24\\
    			& & METIER          & 80.90 $\pm$ 7.51  & 80.74 $\pm$ 8.51  & 72.66 $\pm$ 9.53 \\
    			& & Adversarial CNN & 80.91 $\pm$ 15.39 & 80.79 $\pm$ 16.33 & 73.73 $\pm$ 16.05\\
    			& & TASKED		& \textbf{83.04 $\pm$ 11.35} & \textbf{82.93 $\pm$ 12.35}	& \textbf{75.21 $\pm$ 11.74}\\
    			\hline
    			\multirow{6}{*}{MHEALTH} 
    			& &  MC-CNN			& 85.31 $\pm$ 8.81  & 82.46 $\pm$ 10.50 & 82.33 $\pm$ 10.64\\
    			& & DeepConvLSTM	& 85.46 $\pm$ 7.80  & 83.22 $\pm$ 8.91  & 83.40 $\pm$ 8.68\\
    			& & Self-Attention  & 86.42 $\pm$ 10.41 & 83.85 $\pm$ 12.16 & 84.33 $\pm$ 11.58\\
    			& & METIER          & 87.80 $\pm$ 6.29  & 86.02 $\pm$ 7.59  & 86.70 $\pm$ 7.18\\
    			& & Adversarial CNN & 91.82 $\pm$ 7.54  & 90.53 $\pm$ 8.67  & 91.03 $\pm$ 8.07\\
    			& & TASKED		& \textbf{95.00 $\pm$ 3.72} & \textbf{94.66 $\pm$ 4.14}	& \textbf{94.42 $\pm$ 4.09}\\
    			\hline
    			\multirow{6}{*}{RealDISP} 
    			& & MC-CNN			& 70.19 $\pm$ 15.19 & 70.41 $\pm$ 14.14 & 70.13 $\pm$ 15.01\\
    			& & DeepConvLSTM	& 68.86 $\pm$ 17.03 & 69.74 $\pm$ 15.30 & 68.33 $\pm$ 15.75\\
    			& & Self-Attention  & 85.21 $\pm$ 16.04 & 85.93 $\pm$ 14.15 & 85.41 $\pm$ 15.16\\
    			& & METIER          & 82.39 $\pm$ 15.22 & 83.16 $\pm$ 12.52 & 82.67 $\pm$ 14.05\\
    			& & Adversarial CNN & 78.32 $\pm$ 18.03 & 78.66 $\pm$ 16.72 & 78.65 $\pm$ 16.95\\
    			& & TASKED		& \textbf{86.49 $\pm$ 18.64} & \textbf{87.39 $\pm$ 16.63} & \textbf{87.53 $\pm$ 17.24}\\
    			\hline
    		\end{tabular}}
    	\end{table*}
    	
    	\begin{figure*}[!t]
    	    \centering
    	    \vspace{-0.25cm}
    	    \subfigure[]{
			    \includegraphics[width=0.88\linewidth]{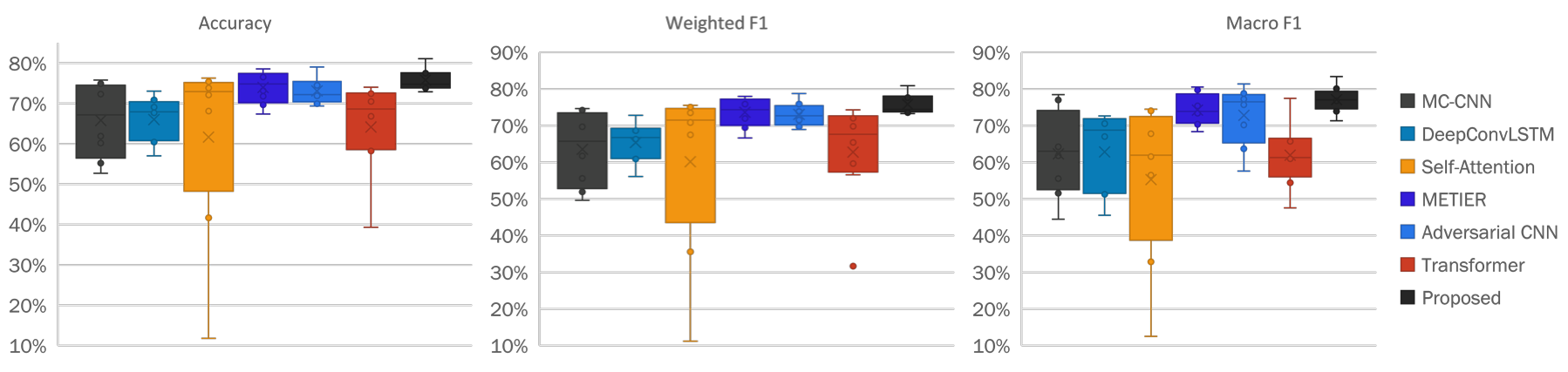}
		    }
		    \vspace{-0.25cm}
		    \subfigure[]{
			    \includegraphics[width=0.88\linewidth]{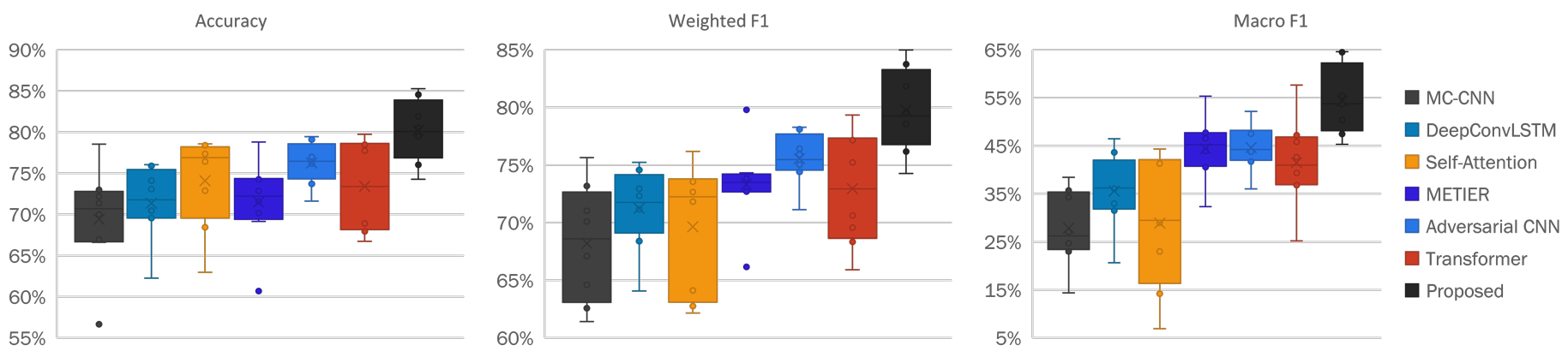}
		    }
		    \vspace{-0.25cm}
		    \subfigure[]{
			    \includegraphics[width=0.88\linewidth]{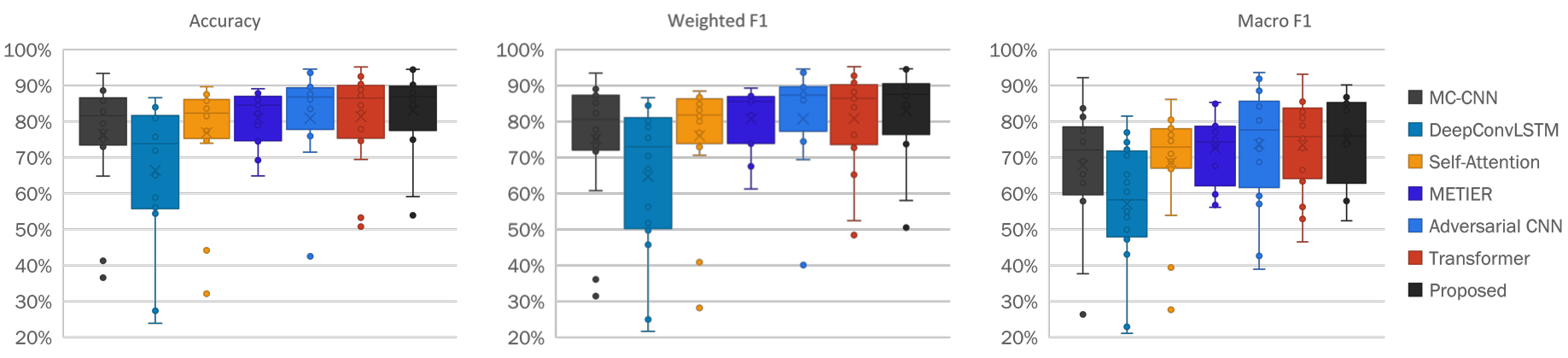}
		    }
		    \vspace{-0.25cm}
		    \subfigure[]{
			    \includegraphics[width=0.88\linewidth]{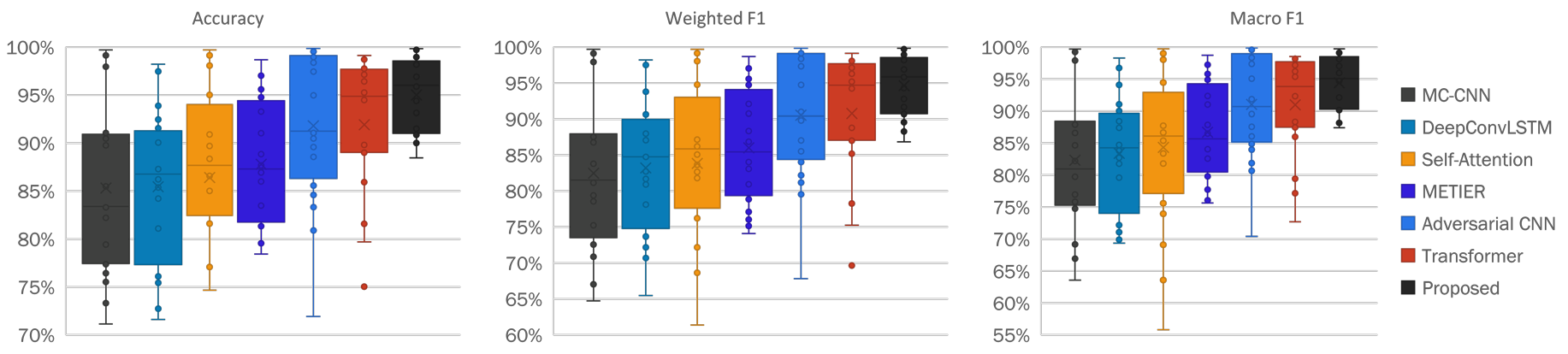}
		    }
		    \vspace{-0.25cm}
		    \subfigure[]{
			    \includegraphics[width=0.88\linewidth]{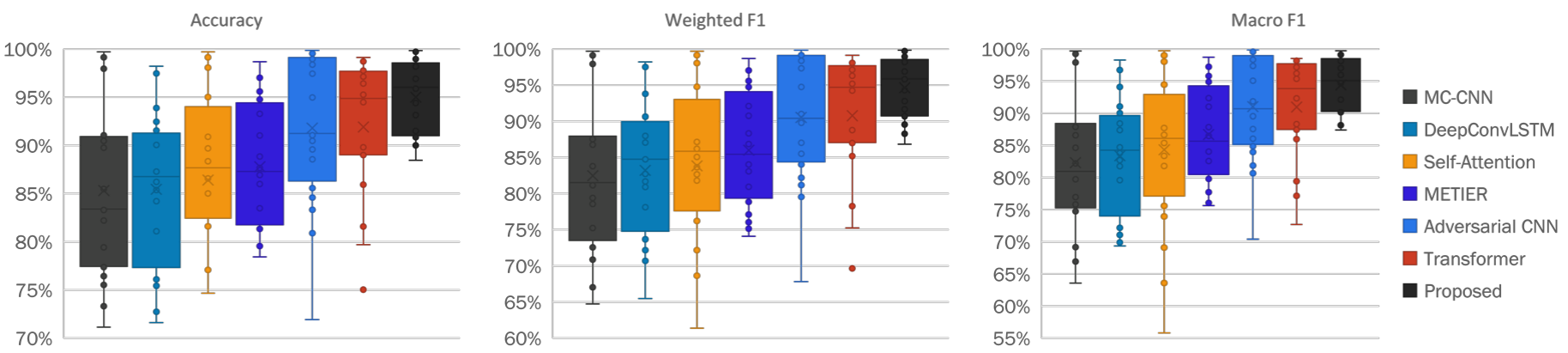}
		    }
		    \vspace{-0.2cm}
    		\caption{Box-and-whisker plots of comparison results with the state-of-the-art in terms of $acc$, $F_w$, and $F_m$ on Opportunity dataset (a) for the mode of locomotion recognition task, (b) for the gesture recognition, (c) PAMAP2 dataset, (d) MHEALTH dataset, and (e) RealDISP dataset.}
    		\label{fig:comparison_boxwhistker}
	    \end{figure*}
	    
	   % \begin{figure*}[!t]
    % 	    \centering
		  %  \subfigure[]{
			 %   \includegraphics[width=0.9\linewidth]{Figures/comparison_PAMAP2}
		  %  }
		  %  \subfigure[]{
			 %   \includegraphics[width=0.9\linewidth]{Figures/comparison_MHEALTH}
		  %  }
		  %  \subfigure[]{
			 %   \includegraphics[width=0.9\linewidth]{Figures/comparison_RealDISP}
		  %  }
    % 		\caption{Box-and-whisker plots of comparison results with the state-of-the-art in terms of $acc$ and $F_m$ on (a) PAMAP2 dataset, (b) MHEALTH dataset, and (c) RealDISP datset.}
    % 		\label{fig:comparison_boxwhistker}
	   % \end{figure*}
	        
    	\cref{tab:comparisonresults} shows the quantitative evaluation results on the Opportunity, PAMAP2, MHEALTH, and RealDISP datasets. Because the Opportunity dataset for the gesture recognition has 18 types of activities and is severely imbalanced, the results in terms of $F_m$ are relatively lower and have a larger standard deviation than the results in terms of $acc$ and $F_w$. The results show that the proposed TASKED framework achieves significantly higher performance on the Opportunity dataset for both locomotion and gesture recognition tasks in terms of all three of the measurements. Furthermore, TASKED provides high-performance results with small inter-subject variation, whereas the state-of-the-arts give high standard deviation results relatively. By comparing the proposed method with the previous method "Adversarial CNN", the proposed transformer architecture and self-knowledge distillation improve the performance. For example, TASKED achieves 4.31, 9.75, 1.48, 3.39, 8.18 percent points improvements over the adversarial CNN in terms of $F_m$ on the Opportunity dataset (for locomotion and gesture tasks), PAMAP2, MHEALTH, and RealDISP datasets, respectively. 

        In \cref{fig:comparison_boxwhistker} (a) and (b), we analyze the results of TASKED and the state-of-the-art methods in the box-and-whisker plots on the Opportunity dataset in terms of $acc$, $F_w$, and $F_m$. The bars signify the performance distribution between 25 \% quantile and 75 \% quantile. The proposed method not only provides significantly better performance, but also gives small performance variances on both tasks than the state-of-the-art methods. 
        
        Additionally, the performance variances on the PAMAP2 and RealDISP datasets are much larger than on the Opportunity and MHEALTH datasets, because of the diversity between different subjects and the data quality. The proposed TASKED outperforms all the state-of-the-art methods on the MHEALTH datasets in terms of all three metrics. TASKED achieves 3.08, 3.89, 3.39 percent points improvements over the best state-of-the-art method in terms of $acc$, $F_w$, and $F_m$, respectively. Additionally, the standard deviation of the performance by TASKED is much smaller than that of the state-of-the-art methods. The comparison results demonstrate the superiority of TASKED. \cref{fig:comparison_boxwhistker} (c)-(e) shows the box-and-whisker plots on the PAMAP2, MHEALTH, and REALDISP datasets in terms of $acc$, $F_w$. and $F_m$. The box-and-whisker plots also show that the proposed method achieves higher minimum performance and the ranges of the performance than other methods.
        
\begin{figure}[!t]
            \centering
            \includegraphics[width=\linewidth]{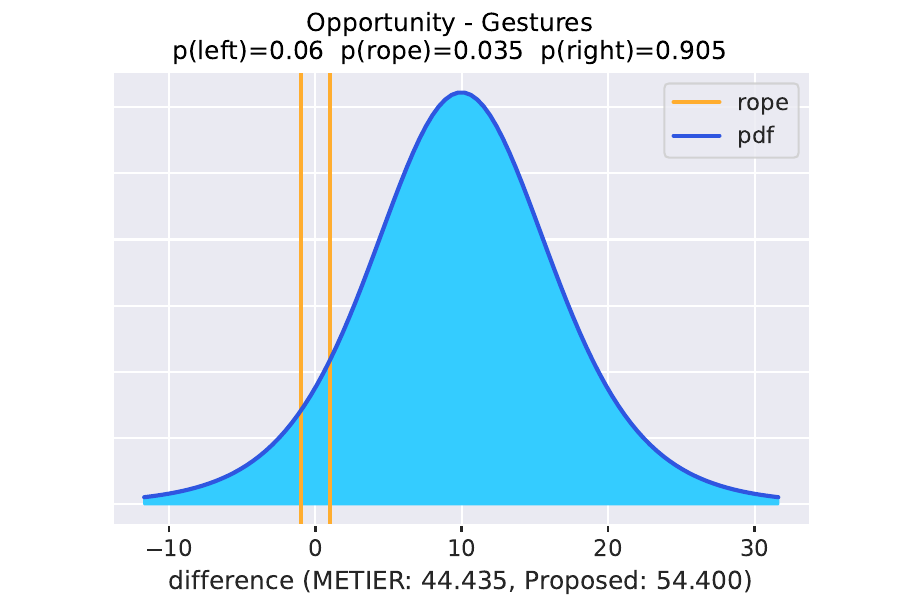}
            \caption{Posterior of the Bayesian correlated t-test for the difference between the macro F1 scores obtained by METIER and our proposed TASKED framework.}
            \label{fig:bayes_comp_single}
        \end{figure}

        It is also important to analyze the statistical significance of the difference between the performance of the classifiers. This can be achieved through frequentist statistics null hypothesis testing, but those can suffer from problems such as p-hacking, but also the more fundamental problem: they do not directly answer the question at hand, that is, their confidence levels are not the probability that a classifier is better than another, but only the probability of getting the observed (or larger) differences assuming the classifiers are equal (null hypotheses) \cite{benavoli2017time}. On the other hand, Bayesian methods, such as the ones proposed in \cite{benavoli2017time} can directly estimate the probability one classifier's performance differs from another by some factor in a single dataset or across different ones.
        In this work we will use their bayesian approach to analyze our results. We use a region of practical equivalence (rope) of 1\% in $F_m$, that is, we want to know the probability that the difference between the model's performance is within 1\% ($p(\text{rope})$), less than that ($p(\text{left})$) or more ($p(\text{right})$).
        For between single datasets, this can be done by the Bayesian correlated t-test \cite{corani2015bayesian} including a correction for underestimation of variance due to overlapping training sets \cite{nadeau1999inference}. For example, \cref{fig:bayes_comp_single} shows the PDF of the posterior of the test for the differences in $F_m$ in the gestures task of the Opportunity dataset. Here we can see that our proposed framework outperforms METIER in that dataset by more than 1\% of $F_m$ with a probability more than 90\%, while there is less than 4\% probability their difference is at the rope and around 6\% that METIER outperforms our method by more than 1\%.
        
        \begin{figure}[!t]
            \centering
            \includegraphics[width=\linewidth]{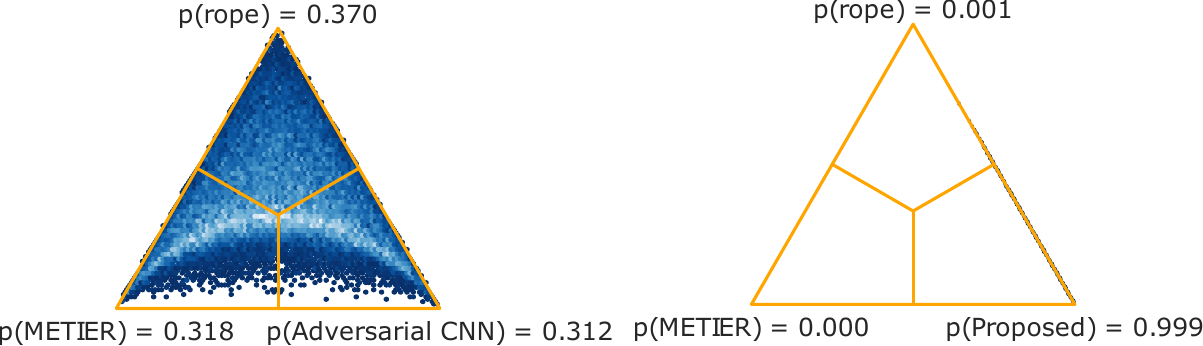}
            \caption{Posterior for METIER versus Adversarial CNN (left) and METIER versus our proposed method (right), both for the sign-rank test.}
            \label{fig:bayes_comp_multi}
        \end{figure}
        As we have tested approaches across different datasets, we can also carry out Bayesian analysis taking into account all tested datasets by using the Bayesian signed rank test \cite{benavoli2014bayesian}. Here we use the same interval for the rope and show the comparison examples in \cref{fig:bayes_comp_multi}. We can see that METIER has likely a comparable performance to the Adversarial CNN, as the $p(\text{rope})$ is higher than both $p(\text{left})$ and $p(\text{right})$. On the other hand, the proposed framework has a 99.9\% probability of being at least 1\% better than METIER, which highlights the improvement obtained by our approach.
        
        \cref{tab:computation} lists the computation complexity and performance of the proposed method with the state-of-the-art methods on PAMAP2 dataset. we evaluate the computation costs with the number of the parameters and the floating-point operations (FLOPs). FLOPs are generally used to represent the execution time or computational complexity of the neural network model \cite{jouppi2017datacenter}. The number of parameters and the FLOPs were computed based on \cite{ptflops}. In addition, we evaluate the performance of the proposed framework with the state-of-the-art methods in terms of the inference time. As shown in \cref{tab:computation}, even though the proposed network architecture has the biggest number of parameters, it provided much faster inference performance than other methods based on LSTM and Transformer such as Deep ConvLSTM, Self-Attention, and METIER. In addition, the TASKED framework provided slower inference time and bigger FLOPs than the CNN-based methods such as MC-CNN and Adversarial CNN. Still, the inference time of the proposed method was less than 1 second for all samples of a subject (test set) and thus it could easily be used for a real-time process.
        \begin{table*}[!t]
            \caption{Comparisons of computational complexity with other methods on PAMAP2 dataset.}
            \label{tab:computation}
            \centering
            \begin{tabular}{llccc}
                \hline
                 Method & Input Dimension   & Parameters (k)  & FLOPs   & Inference Time (s)\\
                 \hline
                 MC-CNN & 36 $\times$ 200           & 273.12    & 2.97M     & 0.1662\\
                 DeepConvLSTM   & 36 $\times$ 200   & 306.32    & 2.85G     & 3.3501\\
                 Self-Attention & 36 $\times$ 200   &  430.41         &     29.40G      &   3.6832 \\
                 METIER & 36 $\times$ 200           & 493.99    & 85.62G    & 2.9548\\
                 Adversarial CNN & 36 $\times$ 200  & 1052.02   & 27.91M    & 0.2067\\
                 TASKED & 36 $\times$ 200           & 1621.37   & 14.80G    & 0.6270\\
                 \hline
            \end{tabular}
        \end{table*}
    \subsection{Ablation Study}
        \begin{table*}[!t]
    		\caption{Ablation study for the proposed framework on Opportunity, PAMAP2, MHEALTH, and RealDISP datasets. The numbers are expressed in percent and represented as $mean \pm std$.}
    		\label{tab:ablationstudy}
    		\centering
    		\renewcommand{\arraystretch}{0.8}
    		\resizebox{\linewidth}{!}{
    		\begin{tabular}{lllccc}
    			\hline
    			Dataset & Task	&Method 			& $acc$				& $F_w$				& $F_m$\\
    			\hline
    			\multirow{8}{*}{Opportunity} &	
    			\multirow{4}{*}{Locomotion} 	
    			& Adversarial CNN & 73.11 $\pm$ 2.98  & 73.06 $\pm$ 3.04  & 72.78 $\pm$ 7.73 \\
    			& & Transformer     & 64.19 $\pm$ 10.97 & 62.80 $\pm$ 13.19 & 61.95 $\pm$ 8.24 \\
    			& & No SelfKD       & 75.78 $\pm$ 2.36 & 75.59 $\pm$ 2.31 & 74.70 $\pm$ 7.04 \\
     			& & TASKED		& \textbf{75.83 $\pm$ 2.54} & \textbf{75.83 $\pm$ 2.57}	& \textbf{77.09 $\pm$ 3.41}\\
    			\cline{2-6}
    			&\multirow{4}{*}{Gestures} 
    			& Adversarial CNN & 76.25 $\pm$ 2.42  & 75.54 $\pm$ 2.13  & 44.65 $\pm$ 4.54 \\
    			& & Transformer     & 73.41 $\pm$ 5.32  & 72.95 $\pm$ 4.62  & 41.50 $\pm$ 8.82\\
    			& & No SelfKD       & 76.95 $\pm$ 2.10 & 76.37 $\pm$ 2.25 & 44.21 $\pm$ 5.92 \\
    			& &TASKED		& \textbf{80.20 $\pm$ 3.55} & \textbf{79.75 $\pm$ 3.40}	& \textbf{54.40 $\pm$ 6.66}\\
    			\hline
    			\multirow{4}{*}{PAMAP2} 
    			& & Adversarial CNN & 80.91 $\pm$ 15.39 & 80.79 $\pm$ 16.33 & 73.73 $\pm$ 16.05\\
    			& & Transformer     & 81.38 $\pm$ 12.93 & 80.81 $\pm$ 13.77 & 73.44 $\pm$ 12.72\\
    			& & No SelfKD     & 81.27 $\pm$ 11.95 & 80.88 $\pm$ 13.48 & 73.45 $\pm$ 12.87\\
    			& & TASKED		& \textbf{83.04 $\pm$ 11.35} & \textbf{82.93 $\pm$ 12.35}	& \textbf{75.21 $\pm$ 11.74}\\
    			\hline
    			\multirow{4}{*}{MHEALTH} 
    			& & Adversarial CNN & 91.82 $\pm$ 7.54  & 90.53 $\pm$ 8.67  & 91.03 $\pm$ 8.07\\
    			& & Transformer     & 91.92 $\pm$ 6.77  & 90.77 $\pm$ 8.29  & 90.97 $\pm$ 7.55\\
    			& & No SelfKD     & 94.79 $\pm$ 4.93  & 94.10 $\pm$ 5.91  & 93.85 $\pm$ 5.62\\
    			& & TASKED		& \textbf{95.00 $\pm$ 3.72} & \textbf{94.66 $\pm$ 4.14}	& \textbf{94.42 $\pm$ 4.09}\\
    			\hline
    			\multirow{4}{*}{RealDISP} 
    			& & Adversarial CNN & 78.32 $\pm$ 18.03 & 78.66 $\pm$ 16.72 & 78.65 $\pm$ 16.95\\
    			& & Transformer     & 83.93 $\pm$ 19.02 & 84.51 $\pm$ 17.69 & 84.96 $\pm$ 18.03\\
    			& & No SelfKD     & 85.54 $\pm$ 18.80 & 86.35 $\pm$ 17.03 & 86.57 $\pm$ 17.46\\
    			& & TASKED		& \textbf{86.49 $\pm$ 18.64} & \textbf{87.39 $\pm$ 16.63} & \textbf{87.53 $\pm$ 17.24}\\
    			\hline
    		\end{tabular}}
    	\end{table*}
    
    To verify the advantages and effectiveness of the proposed TASKED framework architecture, we derive two variants of the proposed method, named ‘Transformer’ and ‘No SelfKD’. ‘Transformer’ is the proposed transformer network alone. This serves as an ablation study, highlighting how much of the improvement comes from the new transformer architecture as opposed to our proposed adversarial framework. ‘No SelfKD’ is to train the proposed TASKED framework using the proposed adversarial learning with MMD regularization but without the pretrained model and self-Knowledge distillation. Because adversarial learning with MMD regularization was already proposed in the previous study, Adversarial CNN, we evaluate the proposed framework without the self-knowledge distillation technique to validate the performance gain of the self-knowledge distillation. In addition, by comparing ‘No SelfKD’ with Adversarial CNN and ‘Transformer’ with other network architectures, we can show the superiority of the proposed transformer architecture.
    
    \cref{tab:ablationstudy} shows the quantitative evaluation results of the ablation study for the proposed transformer architecture with self-knowledge distillation, compared to its two variants and Adversarial CNN, on the Opportunity, PAMAP2, MHEALTH, and RealDISP datasets. Firstly, ‘Transformer’ provided better performance in terms of all three metrics than other network architectures such as MC-CNN, DeepConvLSTM, and Self-Attention. In addition, we confirm the effectiveness of the proposed adversarial learning and MMD regularization by the improved performances of ‘No SelfKD’ from ‘Transformer’. Lastly, it shows that TASKED outperformed its two variants and Adversarial CNN. It means that the self-knowledge distillation method with the adversarial learning scheme improved not only the performance of the activity recognition but also the stability of the training procedure by balancing training optimization between feature generalization and activity recognition.

\begin{figure}
            \centering
            \includegraphics[width=\linewidth]{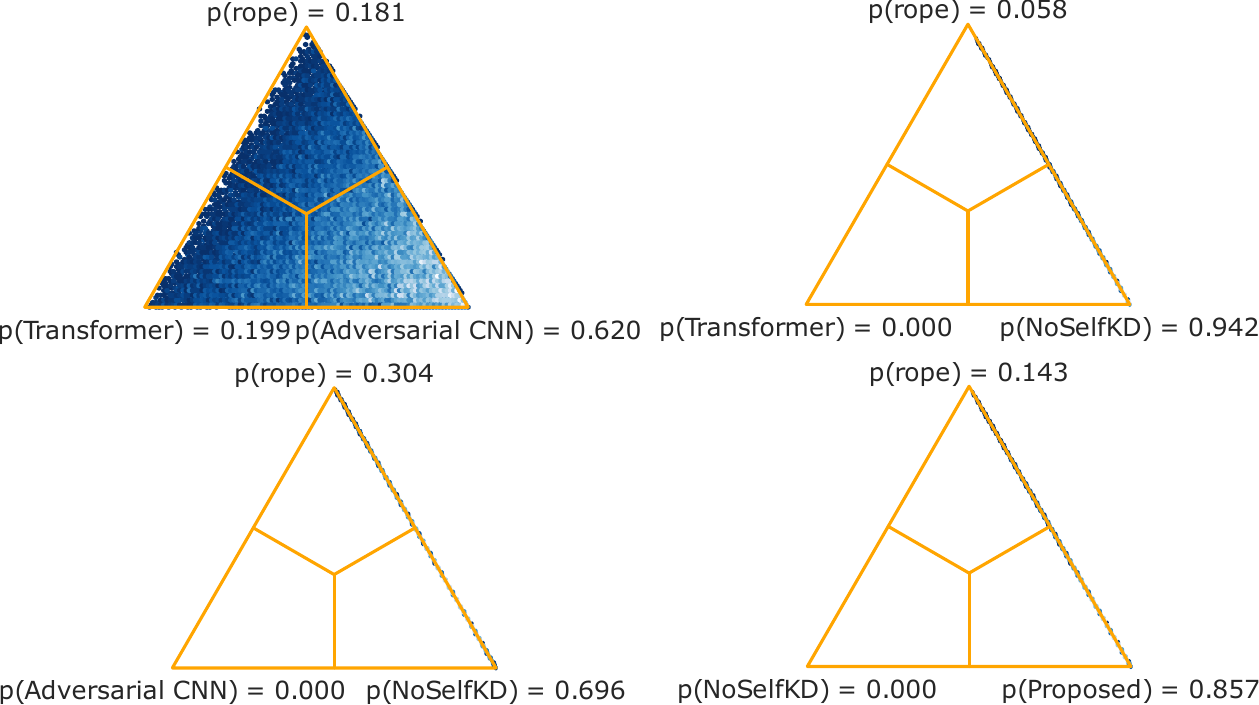}
            \caption{Posterior for the sign-rank test with different comparisons in our ablation study.}
            \label{fig:bayes_cablation_multi}
        \end{figure}

    Regarding statistical comparison as can be seen in \cref{fig:bayes_cablation_multi}, Adversarial CNN is more likely than not to outperform the base Transformer model, showing the importance of adversarial training. The same can be seen when comparing ‘No SelfKD’ with the Adversarial CNN, as in this case, both have adversarial training and the transformer is more likely than not to improve results. That is even clearer when comparing Transformer with ‘No SelfKD’, with over 94\% probability of improvement beyond the rope in the latter case. Some of the performance of TASKED can be attributed with high probability to the knowledge distillation too, as we can see when comparing ‘No SelfKD’ with the proposed framework.
    
	\subsection{Comparison Results across Different Datasets}
	    
	    \begin{table*}[!t]
    		\caption{Comparison results with the state-of-the-art methods across multiple datasets among Opportunity, PAMAP2, MHEALTH, and RealDISP datasets with leave-one-subject-out cross-validation scheme. O, P, M, R denote Opportunity, PAMAP2, MHEALTH, and RealDISP, respectively. The numbers are expressed in percent and represented as $mean \pm std$.}
    		\label{tab:comparisonresultsacross}
    		\centering
    		\resizebox{\linewidth}{!}{
    		\begin{tabular}{cclccc}
    			\hline
    			Dataset & Activities	&Method 			& $acc$				& $F_w$				& $F_m$\\
    			\hline
    			\multirow{6}{*}{O+P+M}    & \multirow{6}{*}{4} 
    			&  MC-CNN			& 79.55 $\pm$ 3.04	& 79.15 $\pm$ 3.19  & 78.78 $\pm$ 5.21\\
    			& & DeepConvLSTM	    & 80.78 $\pm$ 2.00	& 80.59 $\pm$ 2.11	& 79.29 $\pm$ 4.36\\
    			& & METIER            & 83.20 $\pm$ 2.05  & 83.22 $\pm$ 2.04  & 83.92 $\pm$ 2.53 \\
    			& & Adversarial CNN   & 83.01 $\pm$ 1.45  & 83.02 $\pm$ 1.48  & 83.50 $\pm$ 3.05\\
    			& & Transformer       & 82.03 $\pm$ 2.44  & 82.06 $\pm$ 2.48  & 82.94 $\pm$ 2.77\\
    			& & TASKED		    & \textbf{83.74 $\pm$ 2.17} & \textbf{83.76 $\pm$ 2.18}	& \textbf{85.36 $\pm$ 2.16}\\
    			\hline
    			\multirow{6}{*}{O+P+M+R}    & \multirow{6}{*}{13} 
    			&  MC-CNN			    & 76.82 $\pm$ 3.28  & 76.49	$\pm$ 3.56  & 70.71 $\pm$ 6.02\\
    			& & DeepConvLSTM	    & 76.74 $\pm$ 3.77  & 76.92 $\pm$ 3.70  & 70.63 $\pm$ 7.01\\ 
    			& & METIER              & 82.56 $\pm$ 2.27  & 82.55 $\pm$ 2.34  & 81.40 $\pm$ 4.48\\
    			& & Adversarial CNN     & 79.24 $\pm$ 3.91  & 78.72 $\pm$ 4.39  & 77.30 $\pm$ 7.16\\
    			& & Transformer         & 79.84 $\pm$ 2.88  & 79.87 $\pm$ 2.93  & 78.42 $\pm$ 5.77\\
    			& & TASKED		    & \textbf{82.80 $\pm$ 2.09}  & \textbf{82.79 $\pm$ 2.16}  & \textbf{82.82 $\pm$ 4.97}\\
    			\hline
    		\end{tabular}}
        \end{table*}
    
        \begin{figure*}[!t]
    	    \centering
		    \subfigure[]{
			    \includegraphics[width=\linewidth]{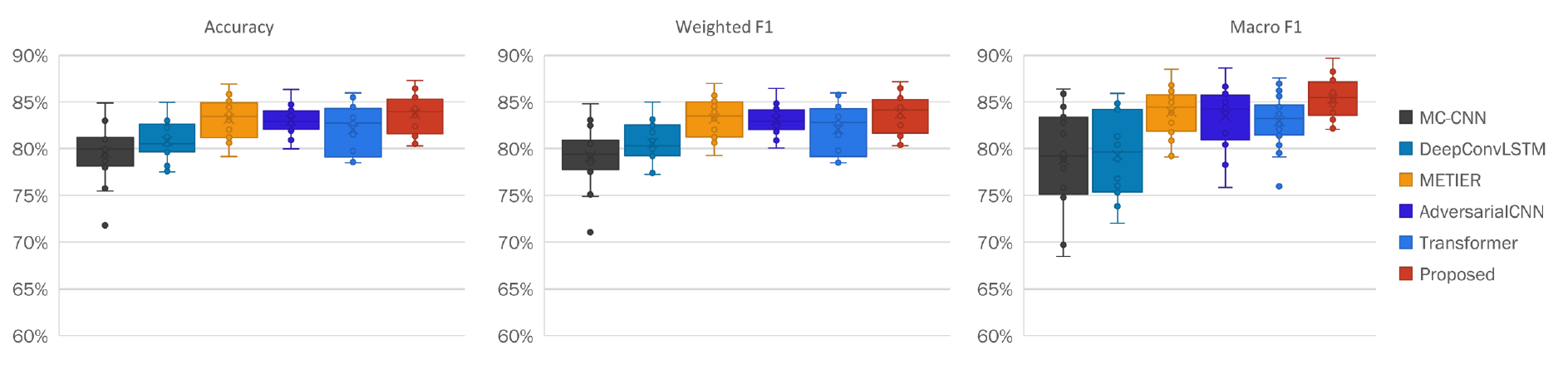}
		    }
		    \subfigure[]{
			    \includegraphics[width=\linewidth]{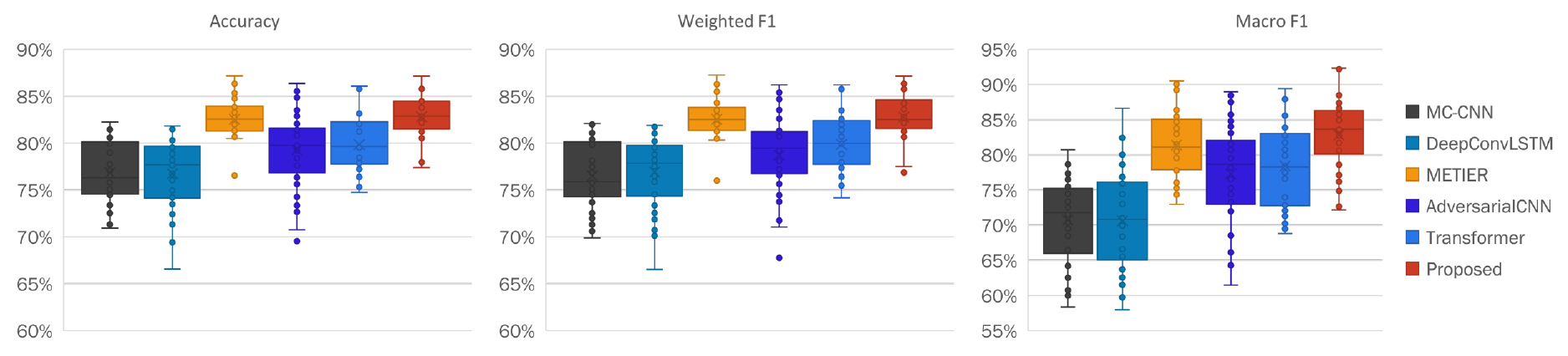}
		    }
    		\caption{Box-and-whisker plots of comparison results with the state-of-the-art methods in terms of $acc$, $F_w$, and $F_m$ across (a) Opportunity, PAMAP2, and MHEALTH datasets with 4 common activities, (b) Opportunity, PAMAP2, MHEALTH, and RealDISP datasets with 13 common activities.}
    		\label{fig:comparison_boxwhistker_across}
	    \end{figure*}
	
	    To show the effectiveness of TASKED further, we performed experiments across multiple datasets with the leave-one-subject-out cross-validation scheme. We have conducted two experiments with two different combinations among the four datasets: Opportunity, PAMAP2, MHEALTH, and RealDISP. Firstly, we extracted common sensor positions and types among four datasets. A total of common sensor data have 18 channels with three sensor positions and three different types. The sensory data is composed of 3 channels of acceleration data acquired from the sensor placed on the subject's back or chest, 9 channels of acceleration, gyroscope, and magnetometer data from the sensor placed on the subject's right hand, and 6 channels of acceleration and gyroscope data from the sensor placed on the subject's left ankle. Since the four datasets have different sampling rates, we resampled all data into $\SI{50}{\hertz}$. We used a sliding window size of 100 with a sliding step of 16, which is 2 seconds of the sliding window and 0.5-second step size. All samples were normalized to zero mean and unit variance. The first combination is Opportunity, PAMAP2, and MHEALTH. A total of 4 common activities, such as lying, sitting, standing, and walking, among the three datasets are used in this work. The second combination uses all four datasets. There are 13 activities that exist in common in at least two or more datasets: lying, sitting, standing, walking, running, cycling, jogging, climbing stairs, knee bending, jumping front and back, waist bends forward, the frontal elevation of arms, and rope jumping. We followed a leave-one-subject-out evaluation scheme. In each fold, one subject data from each dataset was used as the test data, another subject data from each dataset was used as the validation data, and the remaining subjects were used as the training data. 
	    
	    In \cref{tab:comparisonresultsacross}, the comparison results are demonstrated with the state-of-the-art methods with two different dataset combination experiments. The results show that TASKED gives better performance with the two different combinations in terms of $acc$, $F_w$, and $F_m$ than other state-of-the-art methods. In addition, TASKED provides smaller inter-subject variation values than other methods. By comparing the proposed method with the Transformer without the subject discriminator and adversarial learning, the proposed adversarial learning and self-knowledge distillation improved the performance in terms of all three of the measurements and reduced the standard deviation values. The proposed method achieved 1.71, 2.96 percent points improvements over the Transformer in terms of $acc$ and 2.42, 4.40 percent points improvements over the Transformer in terms of $F_m$ with both combinations, respectively. Even though the Transformer was trained without adversarial learning and multi-task learning, the performance of the Transformer provided significantly higher performance than MC-CNN and DeepConvLSTM and is close to the previous method "Adversarial CNN" and the multi-task learning method METIER. It shows that the proposed transformer architecture successfully extracted the spatio-temporal representations from the time-series sensor data. 
	    
	    \cref{fig:comparison_boxwhistker_across} shows the box-and-whisker plots of comparison results with the state-of-the-art in terms of $acc$, $F_w$, and $F_m$ on the two different dataset combinations. The proposed TASKED framework provides the best performance with the smallest variance in terms of all three metrics. 
    
\begin{figure}[!t]
            \centering
            \includegraphics[width=\linewidth]{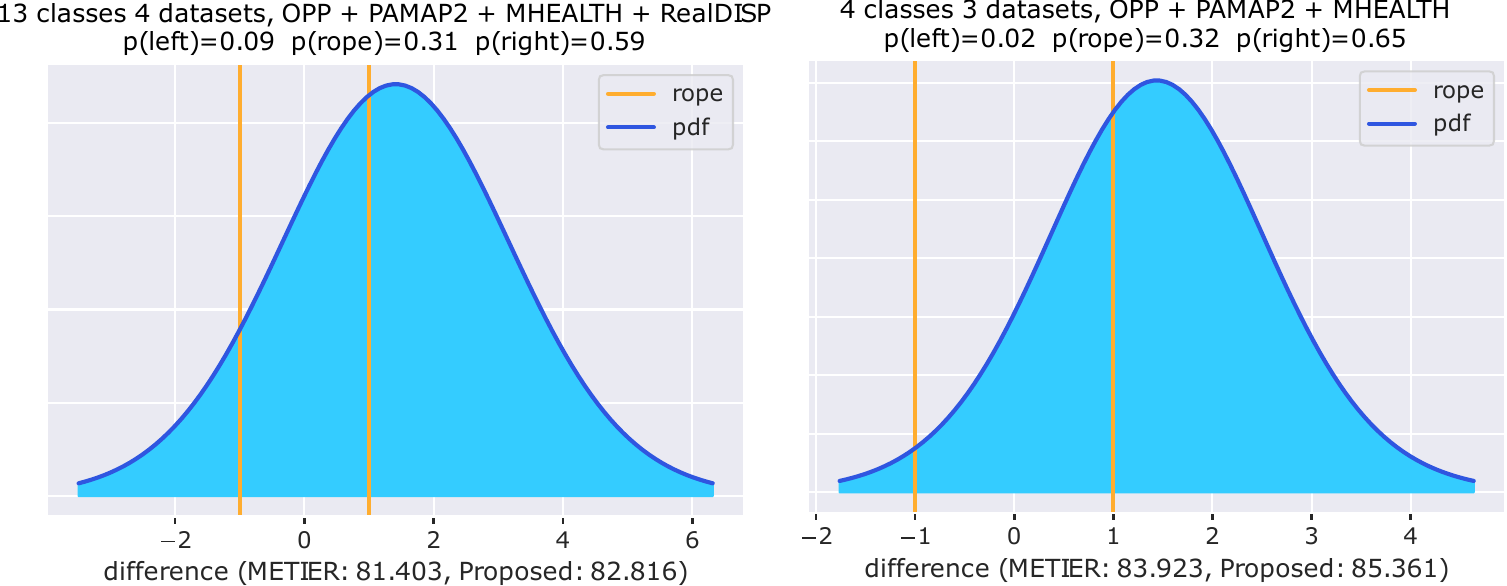}
            \caption{Posterior of the Bayesian correlated t-test for the difference between the macro F1 scores obtained by METIER and our proposed model for test between datasets.}
            \label{fig:bayes_between_single}
        \end{figure}

        Regarding statistical comparison, it is interesting to look closer at the difference in performance between TASKED and METIER. As shown in \cref{fig:bayes_between_single}, in both tests it is more likely than not that TASKED outperforms METIER by more than 1\% (our rope). The probability of them being equivalent is around 30\% for both cases and the probability of METIER being 1\% better are always below 10\%. Even if TASKED achieves a better macro F1 overall and is more likely than not to provide at least 1\% improvement in $F_m$, this gives a better idea of the difference in performance.

	\section{Conclusion}
	\label{sec:conclusion}
	
	In this paper, we have presented a novel TASKED framework for sensor-based HAR. The main objective of our model was to learn cross-domain feature representations by joint optimizing adversarial learning between the transformer-based feature extractor and subject discriminator. The transformer network architecture was designed to learn the spatial and temporal representations from the time-series sensor data, and the adversarial learning scheme with an MK-MMD-based regularization method generalized the feature distributions between different subject domains. In addition, the teacher-free self-knowledge distillation method was adopted to improve the stability of the training procedure and prevent a bias to feature generalization regularized by the MMD loss and adversarial learning. The experiments demonstrated that the proposed method can extract the spatial and temporal representations for HAR from various types of sensor data and generalize the representations among different subject domains. Especially, the average performance in terms of $F_m$ by the proposed method was significantly higher than the results by the state-of-the-art methods on the Opportunity, PAMAP2, MHEALTH, and RealDISP datasets. In summary, the proposed method improved 3.08, 3.89, 3.39 percent points of $acc$, $F_w$, and $F_m$ over the best state-of-the-art method, respectively. Furthermore, the experimental results across different datasets showed that TASKED learns the spatial and temporal representations from different datasets and achieves the best performance among the state-of-the-art methods.
	
	In the future, we plan to deploy the proposed transformer architecture for cross-modal activity recognition, such as wearable sensors and vision. The proposed transformer architecture can learn the spatial and temporal representations from different modalities and transfer knowledge between the modalities. In addition, we can apply the proposed adversarial learning scheme to sensor position-invariant HAR. By using the adversarial scheme, it could recognize the importance of each sensor, generalize the features from different sensors, and optimize their number and placement on the body. The idea is to enable learning across datasets, domains, and modalities.
	
	Furthermore, the proposed method considered each subject a domain, and the MMD loss function was expressed as an average of each pair of subjects. In the case that the number of subjects is large, such as 100 volunteers in practical health /sport monitoring applications, the computational cost of the proposed framework may increase the burden dramatically. In future work, we may try to develop an effective pair sampling method for the average of each pair of subjects or consider adopting contrastive loss instead of the MMD loss.

	\section*{Acknowledgments}
    The research reported in this paper was supported by the BMBF (German Federal Ministry of Education and Research), Germany in the project VidGenSense (01IW21003). It was also funded by the Carl Zeiss Stiftung, Germany under the Sustainable Embedded AI project (P2021-02-009).
	
	\bibliography{mybibfile}

\begin{thebibliography}{10}
\expandafter\ifx\csname url\endcsname\relax
  \def\url#1{\texttt{#1}}\fi
\expandafter\ifx\csname urlprefix\endcsname\relax\def\urlprefix{URL }\fi
\expandafter\ifx\csname href\endcsname\relax
  \def\href#1#2{#2} \def\path#1{#1}\fi

\bibitem{bachlin2009wearable}
M.~Bachlin, M.~Plotnik, D.~Roggen, I.~Maidan, J.~M. Hausdorff, N.~Giladi,
  G.~Troster, Wearable assistant for parkinson’s disease patients with the
  freezing of gait symptom, IEEE Transactions on Information Technology in
  Biomedicine 14~(2) (2009) 436--446.

\bibitem{plotz2012automatic}
T.~Pl{\"o}tz, N.~Y. Hammerla, A.~Rozga, A.~Reavis, N.~Call, G.~D. Abowd,
  Automatic assessment of problem behavior in individuals with developmental
  disabilities, in: Proceedings of the 2012 ACM conference on ubiquitous
  computing, 2012, pp. 391--400.

\bibitem{wen2016adaptive}
J.~Wen, J.~Indulska, M.~Zhong, Adaptive activity learning with dynamically
  available context, in: 2016 IEEE International Conference on Pervasive
  Computing and Communications (PerCom), IEEE, 2016, pp. 1--11.

\bibitem{derawi2010unobtrusive}
M.~O. Derawi, C.~Nickel, P.~Bours, C.~Busch, Unobtrusive user-authentication on
  mobile phones using biometric gait recognition, in: 2010 Sixth International
  Conference on Intelligent Information Hiding and Multimedia Signal
  Processing, IEEE, 2010, pp. 306--311.

\bibitem{direkoglu2012team}
C.~Direkoglu, N.~E. O’Connor, Team activity recognition in sports, in:
  European Conference on Computer Vision, Springer, 2012, pp. 69--83.

\bibitem{sundholm2014smart}
M.~Sundholm, J.~Cheng, B.~Zhou, A.~Sethi, P.~Lukowicz, Smart-mat: Recognizing
  and counting gym exercises with low-cost resistive pressure sensing matrix,
  in: Proceedings of the 2014 ACM international joint conference on pervasive
  and ubiquitous computing, 2014, pp. 373--382.

\bibitem{bao2004activity}
L.~Bao, S.~S. Intille, Activity recognition from user-annotated acceleration
  data, in: International conference on pervasive computing, Springer, 2004,
  pp. 1--17.

\bibitem{chavarriaga2013opportunity}
R.~Chavarriaga, H.~Sagha, A.~Calatroni, S.~T. Digumarti, G.~Tr{\"o}ster,
  J.~d.~R. Mill{\'a}n, D.~Roggen, The opportunity challenge: A benchmark
  database for on-body sensor-based activity recognition, Pattern Recognition
  Letters 34~(15) (2013) 2033--2042.

\bibitem{kwon2018adding}
H.~Kwon, G.~D. Abowd, T.~Pl{\"o}tz, Adding structural characteristics to
  distribution-based accelerometer representations for activity recognition
  using wearables, in: Proceedings of the 2018 ACM international symposium on
  wearable computers, 2018, pp. 72--75.

\bibitem{yang2015deep}
J.~Yang, M.~N. Nguyen, P.~P. San, X.~L. Li, S.~Krishnaswamy, Deep convolutional
  neural networks on multichannel time series for human activity recognition,
  in: Twenty-fourth international joint conference on artificial intelligence,
  2015, p. 3995–4001.

\bibitem{ordonez2016deep}
F.~J. Ord{\'o}{\~n}ez, D.~Roggen, Deep convolutional and lstm recurrent neural
  networks for multimodal wearable activity recognition, Sensors 16~(1) (2016)
  115.

\bibitem{mahmud2020human}
S.~Mahmud, M.~Tonmoy, K.~K. Bhaumik, A.~Rahman, M.~A. Amin, M.~Shoyaib,
  M.~A.~H. Khan, A.~A. Ali, Human activity recognition from wearable sensor
  data using self-attention, arXiv preprint arXiv:2003.09018 (2020).

\bibitem{khaertdinov2021contrastive}
B.~Khaertdinov, E.~Ghaleb, S.~Asteriadis, Contrastive self-supervised learning
  for sensor-based human activity recognition, in: 2021 IEEE International
  Joint Conference on Biometrics (IJCB), IEEE, 2021, pp. 1--8.

\bibitem{cutting1977recognizing}
J.~E. Cutting, L.~T. Kozlowski, Recognizing friends by their walk: Gait
  perception without familiarity cues, Bulletin of the psychonomic society
  9~(5) (1977) 353--356.

\bibitem{singh2017transforming}
M.~S. Singh, V.~Pondenkandath, B.~Zhou, P.~Lukowicz, M.~Liwickit, Transforming
  sensor data to the image domain for deep learning—an application to
  footstep detection, in: 2017 International Joint Conference on Neural
  Networks (IJCNN), IEEE, 2017, pp. 2665--2672.

\bibitem{saputri2014user}
T.~R.~D. Saputri, A.~M. Khan, S.-W. Lee, User-independent activity recognition
  via three-stage ga-based feature selection, International Journal of
  Distributed Sensor Networks 10~(3) (2014) 706287.

\bibitem{hong2015toward}
J.-H. Hong, J.~Ramos, A.~K. Dey, Toward personalized activity recognition
  systems with a semipopulation approach, IEEE Transactions on Human-Machine
  Systems 46~(1) (2015) 101--112.

\bibitem{chen2020metier}
L.~Chen, Y.~Zhang, L.~Peng, Metier: A deep multi-task learning based activity
  and user recognition model using wearable sensors, Proceedings of the ACM on
  Interactive, Mobile, Wearable and Ubiquitous Technologies 4~(1) (2020) 1--18.

\bibitem{sheng2020weakly}
T.~Sheng, M.~Huber, Weakly supervised multi-task representation learning for
  human activity analysis using wearables, Proceedings of the ACM on
  Interactive, Mobile, Wearable and Ubiquitous Technologies 4~(2) (2020) 1--18.

\bibitem{bai2020adversarial}
L.~Bai, L.~Yao, X.~Wang, S.~S. Kanhere, B.~Guo, Z.~Yu, Adversarial multi-view
  networks for activity recognition, Proceedings of the ACM on Interactive,
  Mobile, Wearable and Ubiquitous Technologies 4~(2) (2020) 1--22.

\bibitem{arjovsky2017WGAN}
M.~Arjovsky, S.~Chintala, L.~Bottou, Wasserstein generative adversarial
  networks, in: International Conference on Machine Learning, 2017, pp.
  214--223.

\bibitem{iwasawa2017privacy}
Y.~Iwasawa, K.~Nakayama, I.~Yairi, Y.~Matsuo, Privacy issues regarding the
  application of dnns to activity-recognition using wearables and its
  countermeasures by use of adversarial training., in: IJCAI, 2017, pp.
  1930--1936.

\bibitem{leite2020improving}
C.~F.~S. Leite, Y.~Xiao, Improving cross-subject activity recognition via
  adversarial learning, IEEE Access 8 (2020) 90542--90554.

\bibitem{soleimani2021cross}
E.~Soleimani, E.~Nazerfard, Cross-subject transfer learning in human activity
  recognition systems using generative adversarial networks, Neurocomputing 426
  (2021) 26--34.

\bibitem{gretton2006kernel}
A.~Gretton, K.~Borgwardt, M.~Rasch, B.~Sch{\"o}lkopf, A.~Smola, A kernel method
  for the two-sample-problem, Advances in neural information processing systems
  19 (2006) 513--520.

\bibitem{li2015generative}
Y.~Li, K.~Swersky, R.~Zemel, Generative moment matching networks, in:
  International Conference on Machine Learning, PMLR, 2015, pp. 1718--1727.

\bibitem{long2015learning}
M.~Long, Y.~Cao, J.~Wang, M.~Jordan, Learning transferable features with deep
  adaptation networks, in: International conference on machine learning, PMLR,
  2015, pp. 97--105.

\bibitem{suh2022adversarial}
S.~Suh, V.~F. Rey, P.~Lukowicz, Adversarial deep feature extraction network for
  user independent human activity recognition, in: 2022 IEEE International
  Conference on Pervasive Computing and Communications (PerCom), IEEE, 2022,
  pp. 217--226.

\bibitem{zeng2017semi}
M.~Zeng, T.~Yu, X.~Wang, L.~T. Nguyen, O.~J. Mengshoel, I.~Lane,
  Semi-supervised convolutional neural networks for human activity recognition,
  in: 2017 IEEE International Conference on Big Data (Big Data), IEEE, 2017,
  pp. 522--529.

\bibitem{varamin2018deep}
A.~A. Varamin, E.~Abbasnejad, Q.~Shi, D.~C. Ranasinghe, H.~Rezatofighi, Deep
  auto-set: A deep auto-encoder-set network for activity recognition using
  wearables, in: Proceedings of the 15th EAI International Conference on Mobile
  and Ubiquitous Systems: Computing, Networking and Services, 2018, pp.
  246--253.

\bibitem{dosovitskiy2020image}
A.~Dosovitskiy, L.~Beyer, A.~Kolesnikov, D.~Weissenborn, X.~Zhai,
  T.~Unterthiner, M.~Dehghani, M.~Minderer, G.~Heigold, S.~Gelly, et~al., An
  image is worth 16x16 words: Transformers for image recognition at scale,
  arXiv preprint arXiv:2010.11929 (2020).

\bibitem{plizzari2021skeleton}
C.~Plizzari, M.~Cannici, M.~Matteucci, Skeleton-based action recognition via
  spatial and temporal transformer networks, Computer Vision and Image
  Understanding 208 (2021) 103219.

\bibitem{yuan2020revisiting}
L.~Yuan, F.~E. Tay, G.~Li, T.~Wang, J.~Feng, Revisiting knowledge distillation
  via label smoothing regularization, in: Proceedings of the IEEE/CVF
  Conference on Computer Vision and Pattern Recognition, 2020, pp. 3903--3911.

\bibitem{reiss2012introducing}
A.~Reiss, D.~Stricker, Introducing a new benchmarked dataset for activity
  monitoring, in: 2012 16th international symposium on wearable computers,
  IEEE, 2012, pp. 108--109.

\bibitem{banos2014mhealthdroid}
O.~Banos, R.~Garcia, J.~A. Holgado-Terriza, M.~Damas, H.~Pomares, I.~Rojas,
  A.~Saez, C.~Villalonga, mhealthdroid: a novel framework for agile development
  of mobile health applications, in: International workshop on ambient assisted
  living, Springer, 2014, pp. 91--98.

\bibitem{banos2012benchmark}
O.~Ba{\~n}os, M.~Damas, H.~Pomares, I.~Rojas, M.~A. T{\'o}th, O.~Amft, A
  benchmark dataset to evaluate sensor displacement in activity recognition,
  in: Proceedings of the 2012 ACM Conference on Ubiquitous Computing, 2012, pp.
  1026--1035.

\bibitem{dang2020sensor}
L.~M. Dang, K.~Min, H.~Wang, M.~J. Piran, C.~H. Lee, H.~Moon, Sensor-based and
  vision-based human activity recognition: A comprehensive survey, Pattern
  Recognition 108 (2020) 107561.

\bibitem{bulling2014tutorial}
A.~Bulling, U.~Blanke, B.~Schiele, A tutorial on human activity recognition
  using body-worn inertial sensors, ACM Computing Surveys (CSUR) 46~(3) (2014)
  1--33.

\bibitem{janidarmian2017comprehensive}
M.~Janidarmian, A.~Roshan~Fekr, K.~Radecka, Z.~Zilic, A comprehensive analysis
  on wearable acceleration sensors in human activity recognition, Sensors
  17~(3) (2017) 529.

\bibitem{anguita2012human}
D.~Anguita, A.~Ghio, L.~Oneto, X.~Parra, J.~L. Reyes-Ortiz, Human activity
  recognition on smartphones using a multiclass hardware-friendly support
  vector machine, in: International workshop on ambient assisted living,
  Springer, 2012, pp. 216--223.

\bibitem{hammerla2013preserving}
N.~Y. Hammerla, R.~Kirkham, P.~Andras, T.~Ploetz, On preserving statistical
  characteristics of accelerometry data using their empirical cumulative
  distribution, in: Proceedings of the 2013 international symposium on wearable
  computers, 2013, pp. 65--68.

\bibitem{plotz2011feature}
T.~Pl{\"o}tz, N.~Y. Hammerla, P.~L. Olivier, Feature learning for activity
  recognition in ubiquitous computing, in: Twenty-second international joint
  conference on artificial intelligence, 2011, p. 1729–1734.

\bibitem{lane2015deepear}
N.~D. Lane, P.~Georgiev, L.~Qendro, Deepear: robust smartphone audio sensing in
  unconstrained acoustic environments using deep learning, in: Proceedings of
  the 2015 ACM international joint conference on pervasive and ubiquitous
  computing, 2015, pp. 283--294.

\bibitem{alsheikh2016deep}
M.~A. Alsheikh, A.~Selim, D.~Niyato, L.~Doyle, S.~Lin, H.~P. Tan, Deep activity
  recognition models with triaxial accelerometers, in: 30th AAAI Conference on
  Artificial Intelligence, AAAI 2016, AI Access Foundation, 2016, pp. 8--13.

\bibitem{bhattacharya2016sparsification}
S.~Bhattacharya, N.~D. Lane, Sparsification and separation of deep learning
  layers for constrained resource inference on wearables, in: Proceedings of
  the 14th ACM Conference on Embedded Network Sensor Systems CD-ROM, 2016, pp.
  176--189.

\bibitem{nakano2017effect}
K.~Nakano, B.~Chakraborty, Effect of dynamic feature for human activity
  recognition using smartphone sensors, in: 2017 IEEE 8th International
  Conference on Awareness Science and Technology (iCAST), IEEE, 2017, pp.
  539--543.

\bibitem{hochreiter1997long}
S.~Hochreiter, J.~Schmidhuber, Long short-term memory, Neural computation 9~(8)
  (1997) 1735--1780.

\bibitem{hammerla2016deep}
N.~Y. Hammerla, S.~Halloran, T.~Pl{\"o}tz, Deep, convolutional, and recurrent
  models for human activity recognition using wearables, in: Proceedings of the
  Twenty-Fifth International Joint Conference on Artificial Intelligence, 2016,
  pp. 1533--1540.

\bibitem{guan2017ensembles}
Y.~Guan, T.~Pl{\"o}tz, Ensembles of deep lstm learners for activity recognition
  using wearables, Proceedings of the ACM on Interactive, Mobile, Wearable and
  Ubiquitous Technologies 1~(2) (2017) 1--28.

\bibitem{vaswani2017attention}
A.~Vaswani, N.~Shazeer, N.~Parmar, J.~Uszkoreit, L.~Jones, A.~N. Gomez,
  {\L}.~Kaiser, I.~Polosukhin, Attention is all you need, Advances in neural
  information processing systems 30 (2017).

\bibitem{zeng2018understanding}
M.~Zeng, H.~Gao, T.~Yu, O.~J. Mengshoel, H.~Langseth, I.~Lane, X.~Liu,
  Understanding and improving recurrent networks for human activity recognition
  by continuous attention, in: Proceedings of the 2018 ACM international
  symposium on wearable computers, 2018, pp. 56--63.

\bibitem{morales2016deep}
F.~J.~O. Morales, D.~Roggen, Deep convolutional feature transfer across mobile
  activity recognition domains, sensor modalities and locations, in:
  Proceedings of the 2016 ACM International Symposium on Wearable Computers,
  2016, pp. 92--99.

\bibitem{chiang2017feature}
Y.-T. Chiang, C.-H. Lu, J.~Y.-j. Hsu, A feature-based knowledge transfer
  framework for cross-environment activity recognition toward smart home
  applications, IEEE Transactions on Human-Machine Systems 47~(3) (2017)
  310--322.

\bibitem{handiru2016optimized}
V.~S. Handiru, V.~A. Prasad, Optimized bi-objective eeg channel selection and
  cross-subject generalization with brain--computer interfaces, IEEE
  Transactions on Human-Machine Systems 46~(6) (2016) 777--786.

\bibitem{zhao2020discriminant}
J.~Zhao, L.~Li, F.~Deng, H.~He, J.~Chen, Discriminant geometrical and
  statistical alignment with density peaks for domain adaptation, IEEE
  Transactions on Cybernetics (2020).

\bibitem{cook2013transfer}
D.~Cook, K.~D. Feuz, N.~C. Krishnan, Transfer learning for activity
  recognition: A survey, Knowledge and information systems 36~(3) (2013)
  537--556.

\bibitem{deng2014cross}
W.-Y. Deng, Q.-H. Zheng, Z.-M. Wang, Cross-person activity recognition using
  reduced kernel extreme learning machine, Neural Networks 53 (2014) 1--7.

\bibitem{zhao2011cross}
Z.~Zhao, Y.~Chen, J.~Liu, Z.~Shen, M.~Liu, Cross-people mobile-phone based
  activity recognition, in: Twenty-second international joint conference on
  artificial intelligence, 2011, p. 2545–2550.

\bibitem{wang2018stratified}
J.~Wang, Y.~Chen, L.~Hu, X.~Peng, S.~Y. Philip, Stratified transfer learning
  for cross-domain activity recognition, in: 2018 IEEE international conference
  on pervasive computing and communications (PerCom), IEEE, 2018, pp. 1--10.

\bibitem{khan2018scaling}
M.~A. A.~H. Khan, N.~Roy, A.~Misra, Scaling human activity recognition via deep
  learning-based domain adaptation, in: 2018 IEEE international conference on
  pervasive computing and communications (PerCom), IEEE, 2018, pp. 1--9.

\bibitem{faridee2019augtoact}
A.~Z.~M. Faridee, M.~A. A.~H. Khan, N.~Pathak, N.~Roy, Augtoact: Scaling
  complex human activity recognition with few labels, in: Proceedings of the
  16th EAI International Conference on Mobile and Ubiquitous Systems:
  Computing, Networking and Services, 2019, pp. 162--171.

\bibitem{akbari2019transferring}
A.~Akbari, R.~Jafari, Transferring activity recognition models for new wearable
  sensors with deep generative domain adaptation, in: Proceedings of the 18th
  International Conference on Information Processing in Sensor Networks, 2019,
  pp. 85--96.

\bibitem{zhao2020local}
J.~Zhao, F.~Deng, H.~He, J.~Chen, Local domain adaptation for cross-domain
  activity recognition, IEEE Transactions on Human-Machine Systems 51~(1)
  (2020) 12--21.

\bibitem{wang2018deep}
J.~Wang, V.~W. Zheng, Y.~Chen, M.~Huang, Deep transfer learning for
  cross-domain activity recognition, in: proceedings of the 3rd International
  Conference on Crowd Science and Engineering, 2018, pp. 1--8.

\bibitem{jeyakumar2019sensehar}
J.~V. Jeyakumar, L.~Lai, N.~Suda, M.~Srivastava, Sensehar: a robust virtual
  activity sensor for smartphones and wearables, in: Proceedings of the 17th
  Conference on Embedded Networked Sensor Systems, 2019, pp. 15--28.

\bibitem{goodfellow2014generative}
I.~Goodfellow, J.~Pouget-Abadie, M.~Mirza, B.~Xu, D.~Warde-Farley, S.~Ozair,
  A.~Courville, Y.~Bengio, Generative adversarial nets, in: Advances in neural
  information processing systems, 2014, pp. 2672--2680.

\bibitem{devlin2018bert}
J.~Devlin, M.-W. Chang, K.~Lee, K.~Toutanova, Bert: Pre-training of deep
  bidirectional transformers for language understanding, arXiv preprint
  arXiv:1810.04805 (2018).

\bibitem{liu2020giobalfusion}
S.~Liu, S.~Yao, J.~Li, D.~Liu, T.~Wang, H.~Shao, T.~Abdelzaher, Giobalfusion: A
  global attentional deep learning framework for multisensor information
  fusion, Proceedings of the ACM on Interactive, Mobile, Wearable and
  Ubiquitous Technologies 4~(1) (2020) 1--27.

\bibitem{miao2022towards}
S.~Miao, L.~Chen, R.~Hu, Y.~Luo, Towards a dynamic inter-sensor correlations
  learning framework for multi-sensor-based wearable human activity
  recognition, Proceedings of the ACM on Interactive, Mobile, Wearable and
  Ubiquitous Technologies 6~(3) (2022) 1--25.

\bibitem{wang2019r}
Z.~Wang, Y.~Ma, Z.~Liu, J.~Tang, R-transformer: Recurrent neural network
  enhanced transformer, arXiv preprint arXiv:1907.05572 (2019).

\bibitem{wang2021translating}
Z.~Wang, J.-C. Liu, Translating math formula images to latex sequences using
  deep neural networks with sequence-level training, International Journal on
  Document Analysis and Recognition (IJDAR) 24~(1) (2021) 63--75.

\bibitem{hinton2015distilling}
G.~Hinton, O.~Vinyals, J.~Dean, et~al., Distilling the knowledge in a neural
  network, arXiv preprint arXiv:1503.02531 2~(7) (2015).

\bibitem{nguyen2015recognizing}
L.~T. Nguyen, M.~Zeng, P.~Tague, J.~Zhang, Recognizing new activities with
  limited training data, in: Proceedings of the 2015 ACM International
  Symposium on Wearable Computers, 2015, pp. 67--74.

\bibitem{kingma2014adam}
D.~P. Kingma, J.~Ba, Adam: A method for stochastic optimization, arXiv preprint
  arXiv:1412.6980 (2014).

\bibitem{benavoli2017time}
A.~Benavoli, G.~Corani, J.~Dem{\v{s}}ar, M.~Zaffalon, Time for a change: a
  tutorial for comparing multiple classifiers through bayesian analysis, The
  Journal of Machine Learning Research 18~(1) (2017) 2653--2688.

\bibitem{corani2015bayesian}
G.~Corani, A.~Benavoli, A bayesian approach for comparing cross-validated
  algorithms on multiple data sets, Machine Learning 100~(2) (2015) 285--304.

\bibitem{nadeau1999inference}
C.~Nadeau, Y.~Bengio, Inference for the generalization error, Advances in
  neural information processing systems 12 (1999).

\bibitem{benavoli2014bayesian}
A.~Benavoli, G.~Corani, F.~Mangili, M.~Zaffalon, F.~Ruggeri, A bayesian
  wilcoxon signed-rank test based on the dirichlet process, in: International
  conference on machine learning, PMLR, 2014, pp. 1026--1034.

\bibitem{jouppi2017datacenter}
N.~P. Jouppi, C.~Young, N.~Patil, D.~Patterson, G.~Agrawal, R.~Bajwa, S.~Bates,
  S.~Bhatia, N.~Boden, A.~Borchers, et~al., In-datacenter performance analysis
  of a tensor processing unit, in: Proceedings of the 44th annual international
  symposium on computer architecture, 2017, pp. 1--12.

\bibitem{ptflops}
V.~Sovrasov, \href{https://github.com/sovrasov/flops-counter.pytorch/}{Flops
  counter for convolutional networks in pytorch framework} (2019).
\newline\urlprefix\url{https://github.com/sovrasov/flops-counter.pytorch/}

\end{thebibliography}

\end{document}